\documentclass[5p, times]{elsarticle}

\journal{Journal of \LaTeX\ Templates}

\usepackage{bm}
\usepackage[svgnames,dvipsnames,table]{xcolor}
\usepackage[colorlinks]{hyperref}
\usepackage{amsmath}
\usepackage{amssymb}
\usepackage{adjustbox}
\usepackage{multirow}
\usepackage{tikz}
\usepackage{pgfplots}
\usepgfplotslibrary{patchplots}
\usetikzlibrary{trees,shapes,snakes,calc,matrix,positioning,fit,arrows,patterns,backgrounds,quotes,arrows.meta}
\usepackage{siunitx}
\usepackage{multicol}
\usepackage{caption}
\usepackage{subcaption} 
\usetikzlibrary{shapes.geometric}
\usetikzlibrary{shapes.arrows}
\usepgfplotslibrary{fillbetween}
\usepackage{array}
\usepackage{tabularx}
\usepackage{lipsum} 
 \usepackage{collcell}
\usepackage{hhline}
\usetikzlibrary{spy, calc}
\usepackage{array}
\usepackage{tabularx}
\usepackage{tabulary}
\usepackage{xspace}

\usepackage{bbding}
\usepackage{pifont}
\usepackage{diagbox}
\usepackage{booktabs}
\usepackage{makecell}
\usetikzlibrary{fit}
\usepackage{rotating}
\usepgfplotslibrary{colorbrewer}
\usepackage{fontawesome}
\usepackage{lscape}
\usepackage[para,online,flushleft]{threeparttable}
\usepackage{longtable}
\usepackage{tablefootnote}
\usepackage{makecell}
\usepackage{breakcites}
\usepackage{pifont}
\newcommand{\yes}{\faCheck}%
\newcommand{\no}{\faClose}%

\makeatletter 
\g@addto@macro\TPT@defaults{\footnotesize} 
\makeatother

\newcommand{\colora}{Greens-E}
\newcommand{\colorb}{Oranges-E}
\newcommand{\colorc}{PuOr-G}

\newcommand{\colord}{Blues-G}

\newcommand{\kerneldensity}{2}
\newcommand{\inceptionscore}{3}
\newcommand{\humanobserver}{1}
\newcommand{\classification}{30}
\newcommand{\js}{4}
\newcommand{\mae}{11}
\newcommand{\psnr}{12}
\newcommand{\ssim}{13}
\newcommand{\mi}{{19}}
\newcommand{\vif}{{14}}
\newcommand{\uqi}{{15}}
\newcommand{\fsim}{{17}}
\newcommand{\lpips}{{18}}
\newcommand{\tss}{{29}}
\newcommand{\medianintensity}{{20}}
\newcommand{\lesiondetection}{{31}}
\newcommand{\wasdiscolorhist}{{5}}
\newcommand{\segmentation}{{32}}
\newcommand{\cmr}{{33}}
\newcommand{\qv}{{7}}
\newcommand{\isc}{{8}}
\newcommand{\lineprofile}{{24}}
\newcommand{\noiselevel}{{25}}
\newcommand{\cbr}{{26}}
\newcommand{\suv}{{27}}
\newcommand{\gcf}{{6}}
\newcommand{\depth}{{34}}
\newcommand{\perceptual}{{9}}
\newcommand{\textureloss}{{10}}
\newcommand{\cc}{{21}}
\newcommand{\clinical}{{22}}
\newcommand{\nps}{{28}}
\newcommand{\ifc}{{16}}
\newcommand{\sis}{{23}}

\newcommand{\piglet}{1}
\newcommand{\ldct}{2}
\newcommand{\miccaigcthree}{3}
\newcommand{\lidc}{{4}}
\newcommand{\deeplesion}{{5}}
\newcommand{\wholeheart}{{9}}
\newcommand{\ixi}{11}
\newcommand{\dsbonefive}{12}
\newcommand{\mridata}{13}
\newcommand{\brainles}{15}
\newcommand{\adni}{16}
\newcommand{\bratsfive}{19}
\newcommand{\midas}{{22}}
\newcommand{\bratssix}{{20}}
\newcommand{\bratsseven}{{21}}
\newcommand{\blsa}{{23}}
\newcommand{\isicseven}{{28}}
\newcommand{\isicsix}{{27}}
\newcommand{\phtwo}{{30}}
\newcommand{\cbtc}{{35}}
\newcommand{\miccaiseven}{{36}}
\newcommand{\plco}{{34}}
\newcommand{\montgomery}{{32}}
\newcommand{\jsrt}{{33}}
\newcommand{\humanc}{{24}}
\newcommand{\mitos}{{37}}
\newcommand{\glas}{{38}}
\newcommand{\ochd}{{39}}
\newcommand{\camelyon}{{40}}
\newcommand{\histstain}{{41}}
\newcommand{\drive}{{43}}
\newcommand{\stare}{{44}}
\newcommand{\hrf}{{45}}
\newcommand{\messidor}{{46}}
\newcommand{\dridb}{{47}}
\newcommand{\diaretdb}{{48}}
\newcommand{\ivus}{{52}}
\newcommand{\bratsthree}{{18}}
\newcommand{\wholeheartvessel}{10}
\newcommand{\miccaionetwo}{17}
\newcommand{\rimone}{49}
\newcommand{\ithreea}{50}
\newcommand{\mivia}{51}
\newcommand{\dermofit}{31}
\newcommand{\inbreast}{53}
\newcommand{\ddsm}{54}
\newcommand{\celldetect}{42}
\newcommand{\lits}{6}
\newcommand{\iseg}{25}
\newcommand{\spinebiomedia}{7}
\newcommand{\isiceight}{29}
\newcommand{\nsclc}{8}
\newcommand{\ski}{14}
\newcommand{\biobank}{26}

\newcommand{\ladv}{1}
\newcommand{\limage}{2}
\newcommand{\lcycle}{3}
\newcommand{\lgrad}{4}
\newcommand{\ledge}{5}
\newcommand{\lsharp}{6}
\newcommand{\lshape}{7}
\newcommand{\lpercep}{8}
\newcommand{\lstructure}{9}
\newcommand{\lstylecontent}{11}
\newcommand{\lselfreg}{12}
\newcommand{\lsteer}{13}
\newcommand{\lclassify}{14}
\newcommand{\lfreq}{15}
\newcommand{\lkl}{16}
\newcommand{\lsaliency}{17}
\newcommand{\lphysical}{18}
\newcommand{\lreg}{19}
\newcommand{\lstructuretwo}{10}

\newcommand{\ML}{\mathcal{L}}

 \tikzset{
 myarrow/.style={
 ->,>=stealth,thick
 }
 }
 
  \tikzset{
 myobj/.style={
  rectangle, thick, draw, minimum size=0.6cm,inner sep=0
 }
 }
 
 \tikzset{
 myimage/.style={
  circle, thick, draw, minimum size=0.6cm,inner sep=0
 }
 }

  \tikzset{
 myimagefake/.style={
  circle, thick, draw, minimum size=0.6cm,inner sep=0,dashed
 }
 }
 
   \tikzset{
 domainA/.style={
  draw=black, fill=\colora
 }
 }
   \tikzset{
 domainB/.style={
  draw=black, fill=\colorb
 }
 }
    \tikzset{
 mybk/.style={
  fill=\colorc
 }
 }
 
  \tikzset{
 mybdbox/.style={
  rectangle, thick
 }
 }

  \tikzset{
 mybdboxdashed/.style={
  mybdbox, dashed
 }
 }
 \tikzset{
 mydecoder/.style={
trapezium, shape border rotate=180, draw, thick, rounded corners=0.1cm, minimum size=0.5cm,inner sep=0, mybk
 }
 }
 
 \tikzset{
 myencoder/.style={
trapezium, draw, thick, rounded corners=0.1cm, minimum size=0.5cm,inner sep=0,mybk
 }
 }

\makeatletter
\tikzset{pics/named scope code/.style={code={\tikz@fig@mustbenamed%
  \begin{scope}[local bounding box/.expanded=\tikz@fig@name]#1\end{scope}%
}}}
\makeatother

\tikzset{pics/.cd,
  condition/.style={named scope code={
      \node [circle, draw, thick, minimum size=0.6cm] {$c$};
  }}
}

\tikzset{pics/.cd,
  image/.style={named scope code={
      \node [draw, thick] {$x_g$};
  }}
}

\tikzset{pics/.cd,
  generator/.style={named scope code={
      \node (p0) [rectangle,  thick, rounded corners=0.15cm,minimum size=0.6cm] at (0,0) {};
      \node [trapezium, shape border rotate=0, draw, thick, rounded corners=0.1cm, below=-0.3cm of p0,minimum size=0.5cm,mybk] {};
      \node [trapezium, shape border rotate=180, draw, thick, rounded corners=0.1cm, above=-0.3cm of p0,minimum size=0.5cm,mybk] {};
  }}
}

\tikzset{pics/.cd,
  discriminator/.style={named scope code={
      \node (p0) [trapezium,  draw, thick, rounded corners=0.1cm,minimum size=0.5cm] at (0,0) {D};
  }}
}

\tikzset{ 
    layer/.style={
    rectangle,fill=black!25,minimum height=20pt,inner sep=0pt, outer sep=0pt,minimum width=3pt
    }
 }
 
\tikzset{
	net/.pic={
	\node[layer,fill=gray,rounded corners=1pt](l1){};
	\node[layer,fill=gray,rounded corners=1pt,left= 3pt of l1](l2){};
	\node[layer,fill=gray,rounded corners=1pt,right= 3pt of l1](l3){};
	\node[layer,fill=gray,rounded corners=1pt,right= 3pt of l3](l4){};
	\begin{scope}[on background layer]
	\draw[thick,->,>=stealth]($(l2.west)+(-10pt,0)$) -- ($(l4.east)+(10pt,0)$);
	\end{scope}
	}

}

\newcommand{\parapp}[4]{%
\fill[#4,opacity=.5] (0,0,0)-- (#1,0,0) -- (#1,#3,0)  -- (0,#3,0) --cycle;
\fill[#4,opacity=.5] (0,0,#2)-- (#1,0,#2) -- (#1,#3,#2)  -- (0,#3,#2) --cycle;
\fill[#4,opacity=.5] (0,#3,0)-- (0,#3,#2) -- (#1,#3,#2) -- (#1,#3,0)--cycle;
\fill[#4,opacity=.5] (0,0,0)-- (0,0,#2) -- (#1,0,#2) -- (#1,0,0)--cycle; 
\draw[] (0,0,#2) -- (#1,0,#2) -- (#1,#3,#2) --(0,#3,#2) --(0,0,#2)
        (#1,0,#2) -- (#1,0,0)  -- (#1,#3,0) --(0,#3,0) -- (0,#3,#2)    
        (#1,#3,#2) -- (#1,#3,0);
\draw[dashed,opacity=0.4] (0,0,0) -- (0,0,#2) (0,0,0)-- (#1,0,0) (0,0,0)-- (0,#3,0);

}  

\newcommand{\ca}{blue!10}

\begin{document}
\begin{frontmatter}

\title{Generative Adversarial Network in Medical Imaging: A Review}


\author[1]{Xin Yi\corref{cor1}}
\cortext[cor1]{Corresponding author}
\ead{xiy525@mail.usask.ca}
\author[1,2]{Ekta Walia}
\ead{ewb178@mail.usask.ca}
\author[1]{Paul Babyn}
\ead{Paul.Babyn
@saskhealthauthority.ca}

\address[1]{Department of Medical Imaging, University of Saskatchewan, 103 Hospital Dr, Saskatoon, SK, S7N 0W8 Canada}
\address[2]{Philips Canada, 281 Hillmount Road, Markham, Ontario, ON L6C 2S3, Canada}

\begin{abstract}
Generative adversarial networks have gained a lot of attention in the computer vision community due to their capability of data generation without explicitly modelling the probability density function. The adversarial loss brought by the discriminator provides a clever way of incorporating unlabeled samples into training and imposing higher order consistency. This has proven to be useful in many cases, such as domain adaptation, data augmentation, and image-to-image translation. These properties have attracted researchers in the medical imaging community, and we have seen rapid adoption in many traditional and novel applications, such as image reconstruction, segmentation, detection, classification, and cross-modality synthesis. Based on our observations, this trend will continue and we therefore conducted a review of recent advances in medical imaging using the adversarial training scheme with the hope of benefiting researchers interested in this technique.
\end{abstract}


\end{frontmatter}


\section{Introduction}
With the resurgence of deep learning in  computer vision starting from 2012~\citep{krizhevsky2012imagenet}, the adoption of deep learning methods in medical imaging has increased dramatically. It is estimated that there were over 400 papers published in 2016 and 2017  in major medical imaging related conference venues and journals~\citep{litjens2017survey}. The wide adoption of deep learning in the medical imaging community is due to its demonstrated potential to complement image interpretation and augment image representation and classification.   In this article, we focus on one of the most interesting  recent breakthroughs in the field of deep learning - generative adversarial networks (GANs) - and their potential applications in the field of medical imaging. 
 
 
GANs are a special type of neural network model where two networks are trained simultaneously, with one focused on image generation and the other centered on discrimination. The adversarial training scheme has gained attention  in both academia  and industry  due to its usefulness in counteracting domain shift, and effectiveness in generating new image samples.  This model has achieved state-of-the-art performance in many image generation tasks, including text-to-image synthesis~\citep{xu2017attngan}, super-resolution~\citep{ledig2017photo}, and image-to-image translation~\citep{zhu2017unpaired}.

Unlike deep learning which has its roots traced back to  the 1980s~\citep{fukushima1982neocognitron}, the concept of adversarial training is relatively new with significant recent progress~\citep{goodfellow2014generative}.   This paper presents a general overview of GANs, describes their promising applications in medical imaging, and identifies some remaining challenges that need to be solved to enable their successful application in other medical imaging related tasks. 

\begin{figure*}[tbp]
\centering
\begin{minipage}[t][][b]{0.32\textwidth}
\resizebox{\textwidth}{!}{
\begin{tikzpicture}
  \begin{axis}[title  = (a),
   ybar, ymin=0,
    symbolic x coords = {Synthesis, Reconstruction, Segmentation,   Classification, Detection, Registration,   Others},
    nodes near coords,
    every node near coord/.append style={color=black},
    ylabel=Proportion of publications (\%),
    x tick label style={rotate=45,anchor=east},
xtick pos=left,
ytick pos=left
  ]
  \addplot[\colord,fill=\colord] coordinates { (Synthesis, 46) (Reconstruction, 20)  (Segmentation, 17)   (Detection, 3) (Classification, 8) (Registration, 3) (Others, 3) };
  \end{axis}
\end{tikzpicture}
}
\end{minipage}
\begin{minipage}[t][][b]{0.32\textwidth}
\resizebox{\textwidth}{!}{
\begin{tikzpicture}
  \begin{axis}[title  = (b),
   ybar, ymin=0,
    symbolic x coords = {MR, CT, Histopathology, Retinal Fundus Imaging,   X-ray, Ultrasound,  Dermoscopy, PET,   Mammogram,        Others},
    nodes near coords,
    every node near coord/.append style={color=black},
    ylabel=Proportion of publications (\%),
    x tick label style={rotate=45,anchor=east},
    xtick pos=left,
ytick pos=left,
xtick distance=1,
  ]
  \addplot[\colord,fill=\colord] coordinates { (MR, 43) (CT, 20) (Histopathology, 9)  (PET, 2)  (Ultrasound, 4) (Mammogram, 1) (Dermoscopy, 3) (Retinal Fundus Imaging, 8) (X-ray, 8) (Others, 2) };
  \end{axis}
\end{tikzpicture}
}
\end{minipage}
\begin{minipage}[t][][b]{0.32\textwidth}
\resizebox{\textwidth}{!}{
\begin{tikzpicture}
  \begin{axis}[title  = (c),
   ybar, ymin=0,
    symbolic x coords = {2016, 2017, 2018},
    nodes near coords,
    every node near coord/.append style={color=black},
    ylabel=Number of publications,
    x tick label style={rotate=45,anchor=east},
    xtick distance=1,
    xtick pos=left,
ytick pos=left
  ]
  \addplot[\colord,fill=\colord] coordinates { (2016, 1) (2017, 44)  (2018, 105)};
  \end{axis}
\end{tikzpicture}
}
\end{minipage}
\caption{(a) Categorization of GAN related papers   according to canonical tasks. (b) Categorization of GAN related papers   according to imaging modality. (c) Number of GAN related papers published from 2014. Note that some works performed various tasks and conducted evaluation on datasets with different modalities. We counted these works multiple times in plotting these graphs. Works related to cross domain image transfer were counted based on the source domain. The statistics presented in figure (a) and (b) are based on  papers published on or before January 1st, 2019.}
\label{stat}

\end{figure*}
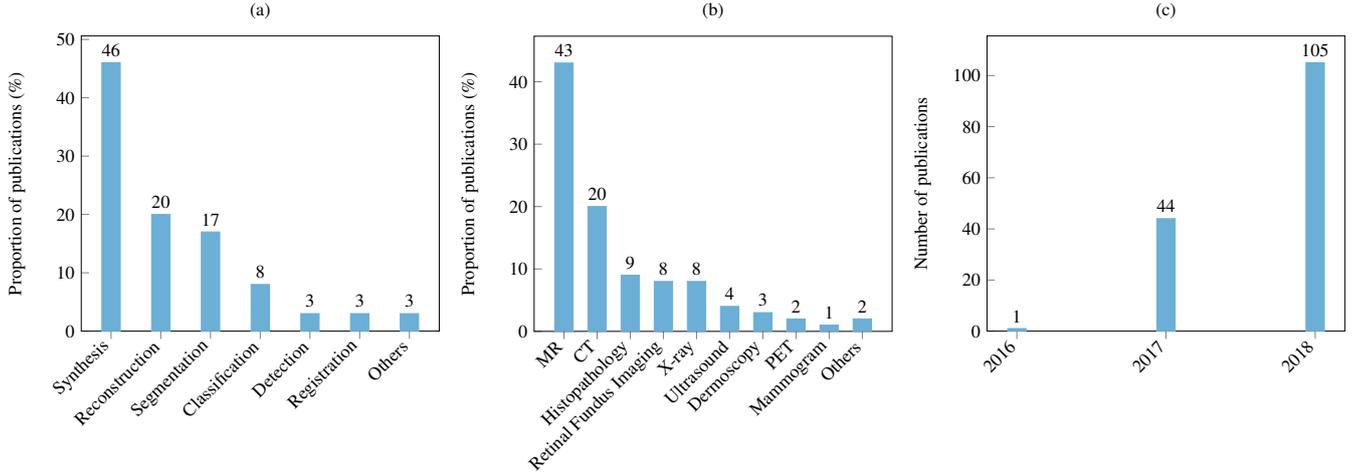

To present a comprehensive overview of all relevant works on GANs in medical imaging, we searched databases including PubMed, arXiv, proceedings of the International Conference  on Medical Image Computing and Computer Assisted Intervention (MICCAI), SPIE Medical Imaging, IEEE International Symposium on Biomedical Imaging (ISBI), and International conference on
Medical Imaging with Deep Learning (MIDL). We also incorporated cross referenced works  not identified  in the above search process. Since there are research publications coming out every month, without losing generality, we set the cut off time of the search as  January 1st, 2019. Works on arXiv that  report only preliminary results are excluded from this review. Descriptive statistics of these papers based on task, imaging modality and year can be found in Figure~\ref{stat}.

\tikzstyle{loosely dashdotted}=[dash pattern=on 11pt off 16pt on \the\pgflinewidth off 18pt]
\def\featuresep{0.9cm}
\def\featuresepspecial{7cm}

\def\featureseptwo{-10pt}
\def\featureseptwospecial{3cm}

\def\featuresepthree{1.75cm}
\def\featuresepthreespecial{2cm}

\def\lsep{6cm}

\begin{figure}[tb]
\centering
\begin{tikzpicture}
\begin{scope}[yshift=0cm]
\node[mydecoder, shape border rotate=90] (g) at (0,0) {G};
\node[below=0.2cm of g.south](gg){    \tikz{
    	\pic(e1) at (0,0) {net};
    	}};

\node (z)[myimage, left=1.2cm of g.west,label=above:\scriptsize{$ \sim p(z)$}]{$z$};
\node (ifake)[right=1.2cm of g.east, myimagefake,label=above:\scriptsize{$\sim p_g(x)$}]{$x_g$};
\draw[thick](z.east) to[out=0,in=-180] (g.west);
\draw[thick, myarrow](g.east) to[out=0,in=-180] (ifake.west);

\node (d)[myencoder,shape border rotate=270] at (4,-0.6) {D};
\node[below=0.2cm of d.south](dd){    \tikz{
    	\pic(e1) at (0,0) {net};
    	}};
\node (ireal)[below=0.5cm of ifake.south, myimage,label=below:\scriptsize{$\sim p_r(x)$}]{$x_r $};
\node (y)[right=1.2cm of d.east, myobj, label=above:real or fake]{$y_1$};

\draw[thick](ifake.east) to[out=0,in=-180] (d.180);
\draw[thick](ireal.east) to[out=0,in=-180] (d.180);

\draw[thick, myarrow](d.east) to[out=0,in=-180] (y.west);

\draw[thick, color=gray!50, dashed]($(z.south)+(0,-1.9cm)$) -- ++(8cm,0);
\end{scope}
\begin{scope}[every label/.append style={text=black, font=\small}]
\node (input)[inner sep=0,label=left:$z$, below=2.3cm of z] {\tikz[scale=0.3]\parapp{0.1}{4}{0.1}{\ca}; };  

\node[right=\featuresep of input.center, anchor=center,label=above:](f1){\tikz[scale=0.3]\parapp{0.5}{8}{0.5}{\ca}; };

\node[right=\featuresep of f1.center, anchor=center,label=above:](f2){\tikz[scale=0.3]\parapp{1}{4}{1}{\ca}; };

\node[right=\featuresep of f2.center, anchor=center,label=above:](f3){\tikz[scale=0.3]\parapp{2}{2}{2}{\ca}; };

\node[right=\featuresep of f3.center, anchor=center,label=right:$x_g$,inner sep=0](f4){\includegraphics[width=0.7cm,height=0.7cm]{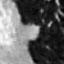}};
\node[anchor=north](or) at(f4.south){or};
\end{scope}
\begin{scope}[xshift=2cm, every label/.append style={text=black, font=\small}]

\node [below=0.5cm of f4,inner sep=0,label=right: $x_r$ ] (input_fake){\includegraphics[width=0.7cm,height=0.7cm]{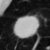}};

\node[right=1.2cm of input_fake.center, yshift=0.6cm, anchor=center,label=above:](d1){\tikz[scale=0.3]\parapp{2}{2}{2}{\ca}; };

\node[right=\featuresep of d1.center, anchor=center,label=above:](d2){\tikz[scale=0.3]\parapp{1}{4}{1}{\ca}; };

\node[right=\featuresep of d2.center, anchor=center,label=above:](d3){\tikz[scale=0.3]\parapp{0.5}{8}{0.5}{\ca}; };

\node[right=\featuresep of d3.center, anchor=center,label=right:](f4){\tikz[scale=0.3]\parapp{0.1}{0.1}{0.1}{\ca}; };  

\end{scope}

\begin{scope}
\node[below right=1.5 and 0 of input.south west, anchor=west](anno_z){\small 
$
\begin{aligned}
z  \in &\text{ } \mathbb{R}^{n\times1\times1}\\
x_g, x_r \in &\text{ } \mathbb{R}^{c\times w\times h}\\
y_1\in &\text{} \{0,1\}
\end{aligned}
$};

\end{scope}
\end{tikzpicture}

\caption{Schematic view of the vanilla GAN for  synthesis of lung nodule  on CT images. Top of the figure shows the network configuration. The part below shows the input, output and the internal feature representations of the generator G and discriminator D. G transforms a sample $z$ from $p(z)$ into a generated nodule $x_g$. D is a binary classifier that differentiates the generated and real  images of lung nodule formed by $x_g$ and $x_r$ respectively.  }
\label{arch}
\end{figure}
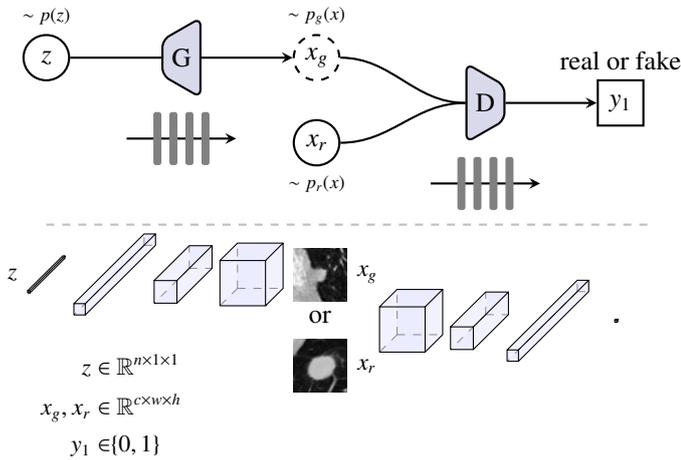

The remainder of the paper is structured as follows.  We begin with a brief introduction of  the principles of  GANs and some of its structural variants in Section~\ref{sect:bg}. It is followed by a comprehensive review of medical image analysis tasks using GANs in Section~\ref{sect:app} including but not limited to the fields of radiology, histopathology and dermatology. We categorize all the works according to canonical tasks: reconstruction, image synthesis, segmentation, classification, detection, registration, and others. Section~\ref{sect:discuss} summarizes the review and discusses prospective applications and identifies  open challenges.

\section{Background}\label{sect:bg}

\subsection{Vanilla GAN}
The vanilla GAN~\citep{goodfellow2014generative} is a generative model that was designed  for  directly drawing samples from the desired data distribution  without the need to explicitly model the underlying  probability density function. It consists of two neural networks: the generator G and the discriminator D. The input to G, $z$ is pure random noise  sampled from a prior distribution $p(z)$, which is commonly chosen to be a Gaussian or a uniform distribution for simplicity. The output of G, $x_g$ is expected to  have visual similarity with the real  sample $x_r$ that is drawn from the real data distribution $p_r(x)$. We denote the non-linear mapping function learned by G parametrized by $\theta_g$   as $x_g = G(z;\theta_g)$.  The input to D is either a real or generated sample. The output of D, $y_1$ is a single value indicating the probability of the input being a real or fake sample. The mapping learned by D parametrized by $\theta_d$ is denoted as $y_1 = D(x;\theta_d)$. The generated samples form a distribution $p_g(x)$ which is desired to be an approximation of $p_r(x)$ after successful training. The top of Figure~\ref{arch} shows an illustration of a vanilla GAN's configuration. G in this example is generating a 2D CT slice depicting a lung nodule.

D's objective is to differentiate these two groups of images whereas the generator G is trained to confuse the discriminator D as much as possible. Intuitively, G could be viewed as a forger trying to produce some quality counterfeit material, and D could be regarded as the police officer trying to detect the forged items. In an alternative view, we can perceive G as receiving a reward signal from D depending upon whether the generated data is accurate or not.  The gradient information is back propagated from D to G, so G adapts its parameters in order to produce an output image that can fool D. The training objectives of D and G can be expressed mathematically as:

\begin{equation}
\begin{split}
\mathcal{L}^{GAN}_{D} &= \max_{D}  \mathbb{E}_{x_r\sim p_r{(x)}}\big[\log D(x_r) \big] 
 + \mathbb{E}_{x_g\sim p_g(x)}\big[\log(1-D(x_g))\big], \\
\mathcal{L}^{GAN}_{G} &= \min_{G}   \mathbb{E}_{x_g\sim p_g(x)}\big[\log(1-D(x_g))\big].
\end{split}
\end{equation}

 As can be seen, D is simply a binary classifier with a maximum log likelihood objective. If  the discriminator D is trained to optimality before the next generator G updates, then minimizing  $\mathcal{L}^{GAN}_{G} $ is proven to be equivalent to minimizing the Jensen--Shannon (JS) divergence between $p_r(x)$ and $p_g(x)$~\citep{goodfellow2014generative}. The desired outcome after training is that samples formed by $x_g$ should approximate the real data distribution $p_r(x)$.


\subsection{Challenges in optimizing GANs}
The above GAN training objective is regarded as a saddle point optimization problem~\citep{yadav2018stabilizing} and the training is  often accomplished by gradient-based methods. G and D are trained alternately from scratch so that they may evolve together. However, there is no guarantee of balance between the training of G and D with the JS divergence. As a consequence,  one network may inevitably  be more powerful than the other, which in most cases  is  D. When D becomes too strong as opposed to G, the generated samples become too easy to be separated from real ones, thus reaching a stage where  gradients from D approach zero, providing no guidance for further training of G.  This happens more frequently when generating high resolution images due to the difficulty of generating meaningful high frequency details.

Another problem commonly faced in training  GANs is mode collapse, which, as the name indicates, is a case when the distribution $p_g(x)$  learned by G focuses on a few limited modes of the data distribution $p_r(x)$. Hence instead of producing  diverse images, it  generates a limited set of samples.

\subsection{Variants of GANs}

\subsubsection{\textbf{Varying objective of D}}\label{sec:objd}
\begin{figure*}[!htbp]
\centering
\resizebox{\textwidth}{!}{
\begin{tikzpicture}
\node(gan)[label=above:{\small{(a) Vanilla GAN}},mybdbox]{
\tikz{
\node[mydecoder] (g) at (0,0) {G};
\node (z)[myimage, below=0.5cm of g.south]{$z$};
\node (ifake)[above=0.5cm of g.north, myimagefake]{$x_g$};
\draw[thick](z.north) to[out=90,in=-90] (g.south);
\draw[thick, myarrow](g.north) to[out=90,in=-90] (ifake.south);

\node (d)[myencoder] at (0.6,2) {D};
\node (ireal)[right=0.5cm of ifake.east, myimage]{$x_r $};
\node (y)[above=0.5cm of d.north, myobj]{$y_1$};

\draw[thick](ifake.north) to[out=90,in=-90] (d.270);
\draw[thick](ireal.north) to[out=90,in=-90] (d.270);

\draw[thick, myarrow](d.north) to[out=90,in=-90] (y.south);

}
};
\node(catgan)[label=above:{\small{(b) CatGAN}},right= of gan.north east, anchor=north west,mybdbox]{
\tikz{
\node[mydecoder] (g) at (0,0) {G};
\node (z)[myimage, below=0.5cm of g.south]{$z$};
\node (ifake)[above=0.5cm of g.north, myimagefake]{$x_g$};
\draw[thick](z.north) to[out=90,in=-90] (g.south);
\draw[thick, myarrow](g.north) to[out=90,in=-90] (ifake.south);

\node (d)[myencoder] at (0.6,2) {D};
\node (ireal)[right=0.5cm of ifake.east, myimage]{$x_r $};
\node (y)[above=0.5cm of d.north, myobj]{$y_2$};

\draw[thick](ifake.north) to[out=90,in=-90] (d.270);
\draw[thick](ireal.north) to[out=90,in=-90] (d.270);

\draw[thick, myarrow](d.north) to[out=90,in=-90] (y.south);

}};

\node(ebgan)[label=above:{\small{(c) EBGAN/BEGAN}},,right= of catgan.north east, anchor=north west, mybdbox]{
\tikz{
\node[mydecoder] (g) at (0,0) {G};
\node (z)[myimage, below=0.5cm of g.south]{$z$};
\node (ifake)[above=0.5cm of g.north, myimagefake]{$x_g$};
\draw[thick](z.north) to[out=90,in=-90] (g.south);
\draw[thick, myarrow](g.north) to[out=90,in=-90] (ifake.south);

\node (de)[myencoder] at (0.6,2) {$\text{D}_e$};
\node (ireal)[right=0.5cm of ifake.east, myimage]{$x_r $};
\node (dd)[above=0.25cm of de.north, mydecoder]{$\text{D}_d$};
\node (y)[above=0.5cm of dd.north, myobj]{$y_3$};

\draw[thick](ifake.north) to[out=90,in=-90] (de.270);
\draw[thick](ireal.north) to[out=90,in=-90] (de.270);

\draw[thick](de.north) to[out=90,in=-90] (dd.south);
\draw[thick, myarrow](dd.north) to[out=90,in=-90] (y.south);

}
};

\node(ali)[label=above:{\small{(d) ALI/BiGAN}},right= of ebgan.north east, anchor=north west,mybdbox]{
\tikz{

\node[mydecoder] (g) at (0,0) {$\text{G}_{d}$ };
\node (z)[myimage, below=0.5cm of g.south]{$z_g$};
\node (ifake)[above =0.5cm of g.north, myimagefake]{$x_g$};
\node (zz)[right =0.2cm of ifake, myimage]{$z_g$};
\node(zzifake)[draw,inner sep=0.1cm, rectangle, fit={(zz) (ifake)}]{};

\draw[thick](z.north) to[out=90,in=-90] (g.south);
\draw[thick, myarrow](g.north) to[out=90,in=-90] (ifake.south);

\node (d0)[myencoder, right= 1.5cm of g] {$\text{G}_{e}$};
\node (zr)[myimage, above=0.5cm of d0.north]{$z_r$};
\node (ir)[myimage, below =0.5cm of d0.south]{$x_r$};

\draw[thick](ir.north)to[out=90,in=-90](d0.south);
\draw[thick,myarrow](d0.north)to[out=90,in=-90](zr.south);

\node (d)[myencoder] at (1.3,2.3) {D};
\node (ireal)[left=0.2cm of zr.west, myimage]{$x_r $};
\node (y)[above=0.5cm of d.north, myobj]{$y_1$};
\node(zrireal)[draw,inner sep=0.1cm, rectangle, fit={(zr) (ireal)}]{};

\draw[thick](zzifake.north) to[out=90,in=-90] (d.270);
\draw[thick](zrireal.north) to[out=90,in=-90] (d.270);

\draw[thick, myarrow](d.north) to[out=90,in=-90] (y.south);

\draw[thick](z.east) to[out=0, in=-90](zz.south);
\draw[thick](ir.west) to[out=-180, in=-90](ireal.south);

}
};

\node(InfoGAN)[label=above:{\small{(e) InfoGAN}},right= 1cm of ali.north east, anchor=north west,mybdbox]{
\tikz{
\node[mydecoder] (g) at (0,0) {G};
\node (z)[myimage, draw, below left=0.5 and 0.1 of g]{$z$};
\node (cin)[right=0.1cm of z, myimage]{$c$};
\node(zc)[draw,inner sep=0.1cm, rectangle, fit={(z) (cin)}]{};

\pic (c)[below right=1.5cm and 0.3cm of g.south]{condition};
\node (ifake)[above=0.5cm of g.north, myimagefake]{$x_g$};
\draw[thick](zc.north) to[out=90,in=-90] (g.270);
\draw[thick, myarrow](g.north) to[out=90,in=-90] (ifake.south);
\draw[thick](c.north) to [out=90,in=-90] (cin.south);

\node (d)[myencoder] at (0.6,2) {D};
\node (ireal)[right=0.5cm of ifake.east, myimage]{$x_r $};

\node (y1)[above left=0.5cm and 0.3cm of d.north, myobj]{$y_1$};
\node (y2)[above right=0.5cm and 0.3cm of d.north, myimage]{c};

\draw[thick](ifake.north) to[out=90,in=-90] (d.270);
\draw[thick](ireal.north) to[out=90,in=-90] (d.270);

\draw[thick, myarrow](d.110) to[out=90,in=-90] (y1.south);
\draw[thick, myarrow](d.70) to[out=90,in=-90] (y2.south);

}
};

\node(acgan)[label=above:{\small{(f) ACGAN}},below =2cm of gan.south west, anchor=north west,mybdbox]{
\tikz{
\node[mydecoder] (g) at (0,0) {G};
\node (z)[myimage, draw, below left=0.5 and 0.1 of g]{$z$};
\node (cin)[right=0.1cm of z, myimage]{$c$};
\node(zc)[draw,inner sep=0.1cm, rectangle, fit={(z) (cin)}]{};

\pic (c)[below right=1.5cm and 0.3cm of g.south]{condition};
\node (ifake)[above=0.5cm of g.north, myimagefake]{$x_g$};
\draw[thick](zc.north) to[out=90,in=-90] (g.270);
\draw[thick, myarrow](g.north) to[out=90,in=-90] (ifake.south);
\draw[thick](c.north) to [out=90,in=-90] (cin.south);

\node (d)[myencoder] at (0.6,2) {D};
\node (ireal)[right=0.5cm of ifake.east, myimage]{$x_r $};

\node (y1)[above left=0.5cm and 0.3cm of d.north, myobj]{$y_1$};
\node (y2)[above right=0.5cm and 0.3cm of d.north, myimage]{c};

\draw[thick](ifake.north) to[out=90,in=-90] (d.270);
\draw[thick](ireal.north) to[out=90,in=-90] (d.270);

\draw[thick, myarrow](d.110) to[out=90,in=-90] (y1.south);
\draw[thick, myarrow](d.70) to[out=90,in=-90] (y2.south);

\draw[thick](c.east) to [out=0,in=-90] (ireal.south);
}
};

\node(vaegan)[label=above:{\small{(g) VAEGAN}},right= of acgan.north east, anchor=north west,mybdbox]{
\tikz{
\node[mydecoder] (g) at (0,0) {$\text{G}_{d}$ };
\node (z)[myimage, below left=0.5cm and 0.5cm of g.south]{$z$};
\node (zr)[myimage, below right=0.5cm and 0.5cm of g.south]{$z_r$};

\node (irrecon)[above right=0.5cm and 0.3cm of g.north, myimagefake]{$\hat{x}_r$};
\node (ifake)[above left=0.5cm and 0.3cm of g.north, myimagefake]{$x_g$};

\draw[thick](z.north) to[out=90,in=-90] (g.south);
\draw[thick](zr.north) to[out=90,in=-90] (g.south);

\draw[thick, myarrow](g.north) to[out=90,in=-90] (ifake.south);
\draw[thick, myarrow](g.north) to[out=90,in=-90] (irrecon.south);

\node (d0)[myencoder, below right=1.5cm and 0.45cm of g.south] {$\text{G}_{e}$};
\node (ir)[myimage, below =0.3cm of d0.south]{$x_r$};
\draw[thick](ir.north)to[out=90,in=-90](d0.south);
\draw[thick,myarrow](d0.north)to[out=90,in=-90](zr.south);

\node (d)[myencoder] at (0.6,2) {D};
\node (ireal)[right=0.3cm of irrecon.east, myimage]{$x_r $};
\node (y)[above=0.5cm of d.north, myobj]{$y_1$};

\draw[thick](ifake.north) to[out=90,in=-90] (d.270);
\draw[thick](ireal.north) to[out=90,in=-90] (d.270);
\draw[thick](irrecon.north) to[out=90,in=-90] (d.270);

\draw[thick, myarrow](d.north) to[out=90,in=-90] (y.south);

}
};

\node(cgan)[label=above:{\small{(h) CGAN}},right= of vaegan.north east, anchor=north west,mybdbox]{
\tikz{
\node[mydecoder] (g) at (0,0) {G};
\node (z)[myimage, draw, below left=0.5 and 0.1 of g]{$z$};
\node (cin)[right=0.1cm of z, myimage]{$c$};
\node(zc)[draw,inner sep=0.1cm, rectangle, fit={(z) (cin)}]{};

\pic (c)[below right=1.5cm and 0.3cm of g.south]{condition};
\node (ifake)[above=0.5cm of g.north, myimagefake]{$x_g$};
\draw[thick](zc.north) to[out=90,in=-90] (g.270);
\draw[thick, myarrow](g.north) to[out=90,in=-90] (ifake.south);
\draw[thick](c.north) to [out=90,in=-90] (cin.south);

\node (d)[myencoder] at (0.6,2) {D};
\node (ireal)[right=0.5cm of ifake.east, myimage]{$x_r $};

\node (y)[above=0.5cm of d.north, myobj]{$y_1$};
\draw[thick](ifake.north) to[out=90,in=-90] (d.270);
\draw[thick](ireal.north) to[out=90,in=-90] (d.270);

\draw[thick, myarrow](d.north) to[out=90,in=-90] (y.south);

\draw[thick](c.east) to[out=0,in=-90] (ireal.south);
}
};

\node(lapgan)[label={[align=center]\small{(i) LAPGAN/SGAN} \\\small{(cascade or stack of GANs)}},mybdbox, right= of cgan.north east, anchor=north west]{
\tikz{
\begin{scope}
\node[mydecoder] (g1) at (0,0) {G1};
\node (z1)[myimage, below=0.5cm of g1.south]{$z$};
\node (i1)[above=0.8cm of g1.north, myimagefake]{$x_{g1}$};
\draw[thick](z1.north) to[out=90,in=-90] (g1.south);
\draw[thick, myarrow](g1.north) to[out=90,in=-90] (i1.south);

\node (d1)[myencoder] at (2.5,0) {D1};
\node (ifake1)[below left=0.5cm and 0.3cm of d1.south, myimagefake]{$x_{g1} $};
\node (ireal1)[below right=0.5cm and 0.3cm of d1.south, myimage]{$x_{r1} $};
\node (y1)[above=0.5cm of d1.north, myobj]{$y_1$};

\draw[thick](ifake1.north) to[out=90,in=-90] (d1.270);
\draw[thick](ireal1.north) to[out=90,in=-90] (d1.270);

\draw[thick, myarrow](d1.north) to[out=90,in=-90] (y1.south);

\draw[thick, dashed](i1.east)  to[out=0,in=-180]  (ifake1.west);
\end{scope}
\begin{scope}[yshift=3cm]
\node[mydecoder] (g2) at (0,0) {G2};
\node (i2)[above=0.8cm of g2.north, myimagefake]{$x_{g2}$};
\draw[thick](i1.north) to[out=90,in=-90] (g2.south);
\draw[thick, myarrow](g2.north) to[out=90,in=-90] (i2.south);

\node (d2)[myencoder] at (2.5,0) {D2};
\node (ifake2)[below left=0.5cm and 0.3cm of d2.south, myimagefake]{$x_{g2} $};
\node (ireal2)[below right=0.5cm and 0.3cm of d2.south, myimage]{$x_{r2} $};
\node (y2)[above=0.5cm of d2.north, myobj]{$y_1$};

\draw[thick](ifake2.north) to[out=90,in=-90] (d2.270);
\draw[thick](ireal2.north) to[out=90,in=-90] (d2.270);

\draw[thick, myarrow](d2.north) to[out=90,in=-90] (y2.south);

\draw[thick, dashed](i2.east)  to[out=0,in=-180]  (ifake2.west);
\end{scope}
\begin{scope}[yshift=6cm]
\node[mydecoder] (g3) at (0,0) {G3};
\node (i3)[above=0.5cm of g3.north, myimagefake]{$x_{g3}$};
\draw[thick](i2.north) to[out=90,in=-90] (g3.south);
\draw[thick, myarrow](g3.north) to[out=90,in=-90] (i3.south);

\node (d3)[myencoder] at (2.5,0) {D3};
\node (ifake3)[below left=0.5cm and 0.3cm of d3.south, myimagefake]{$x_{g3} $};
\node (ireal3)[below right=0.5cm and 0.3cm of d3.south, myimage]{$x_{r3} $};
\node (y3)[above=0.5cm of d3.north, myobj]{$y_1$};

\draw[thick](ifake3.north) to[out=90,in=-90] (d3.270);
\draw[thick](ireal3.north) to[out=90,in=-90] (d3.270);

\draw[thick, myarrow](d3.north) to[out=90,in=-90] (y3.south);

\draw[thick, dashed](i3.east)  to[out=0,in=-180]  (ifake3.west);
\end{scope}

}
};

\node(caption)[below right= 0.5cm and 1cm of acgan.south, anchor=north,mybdbox]{
\tikz{
\node(y1)[label=right:{real or fake sample}, myobj] at (0,0) {$y_1$};
\node(y2)[below= 10pt of y1.south, label=right:{ certain or uncertain class prediction}, myobj]{$y_2$};
\node(y3)[below= 10pt of y2.south, label=right:{ real or fake reconstruction loss}, myobj]{ $y_3$};

}
};

\end{tikzpicture}}
\caption{A schematic view of variants of GAN. $c$ represents the conditional vector. In CGAN and ACGAN, $c$ is the discrete categorical code (e.g. one hot vector) that encodes class labels and in InfoGAN it can also be continuous code that encodes attributes. $x_g$ generally refers to the generated image but can also be internal representations as in SGAN.}
\label{ganvar}
\end{figure*}
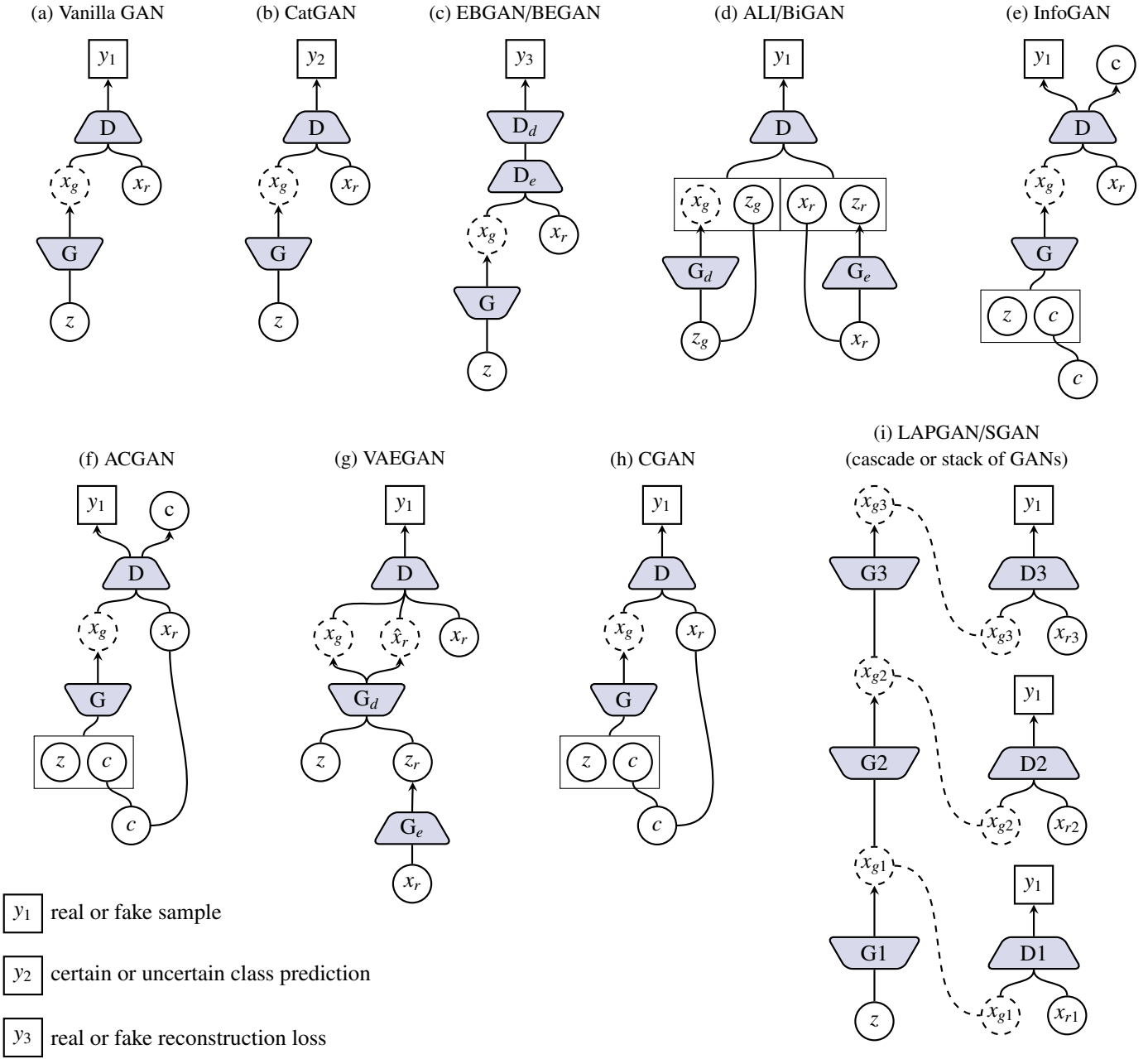

In order to stabilize training and also to  avoid mode collapse, different losses for D have been proposed, such as f-divergence (f-GAN)~\citep{nowozin2016f}, least-square (LSGAN)~\citep{mao2016least}, hinge loss~\citep{miyato2018spectral}, and Wasserstein distance (WGAN, WGAN-GP)~\citep{arjovsky2017wasserstein, gulrajani2017improved}. Among these, Wasserstein distance is arguably the most popular metric. As an alternative to the real/fake discrimination scheme, \cite{springenberg2015unsupervised}  proposed an entropy based objective where real data is encouraged to make confident class predictions (CatGAN, Figure~\ref{ganvar} b).  In EBGAN~\citep{zhao2016energy} and BEGAN~\citep{berthelot2017began} (Figure~\ref{ganvar} c), the commonly used encoder architecture for discriminator is replaced with an autoencoder architecture. D's objective then becomes matching autoencoder loss distribution rather than data distribution. 

GANs themselves lack the mechanism of inferencing the underlying latent vector that is likely to encode the input. Therefore, in ALI~\citep{dumoulin2016adversarially} and BiGAN~\citep{donahue2016adversarial} (Figure~\ref{ganvar} d), a separate encoder network is incorporated. D's objective then becomes separating joint samples $(x_g, z_g)$ and $(x_r, z_r)$. In InfoGAN (Figure~\ref{ganvar} e), the discriminator outputs the latent vector that encodes part of the semantic features of the generated image. The discriminator maximizes the mutual information between the generated image and the latent attribute vector the generated image is conditioned upon. After successful training, InfoGAN can explore inherent data attributes and perform conditional data generation based on these attributes. The use of class labels has been shown to further improve  generated image's quality and this information can be easily incorporated into D by enforcing D to provide class probabilities and use cross entropy loss for optimization such as used in ACGAN~\citep{odena2016conditional} (Figure~\ref{ganvar} f). 

\subsubsection{\textbf{Varying objective of G}}
In the vanilla GAN, G transforms noise $z$ to sample $x_g = G(z)$. This is usually accomplished by using a decoder network to progressively increase the spatial size of the output until the desired resolution is achieved as   shown in Figure~\ref{arch}. ~\cite{larsen2015autoencoding} proposed  a variational autoencoder network (VAE) as the underlying architecture of G (VAEGAN, Figure~\ref{ganvar} g), where it can use pixel-wise reconstruction loss to enforce the decoder part of VAE to generate structures to match the real images.

The original setup of a GAN does not have any restrictions on the modes of data it can generate. However, if auxiliary information were  provided during the generation, the GAN can be driven to output images with desired properties. A GAN in this scenario is usually referred as a conditional GAN (cGAN) and the generation process can be expressed as $x_g = G(z, c)$. 

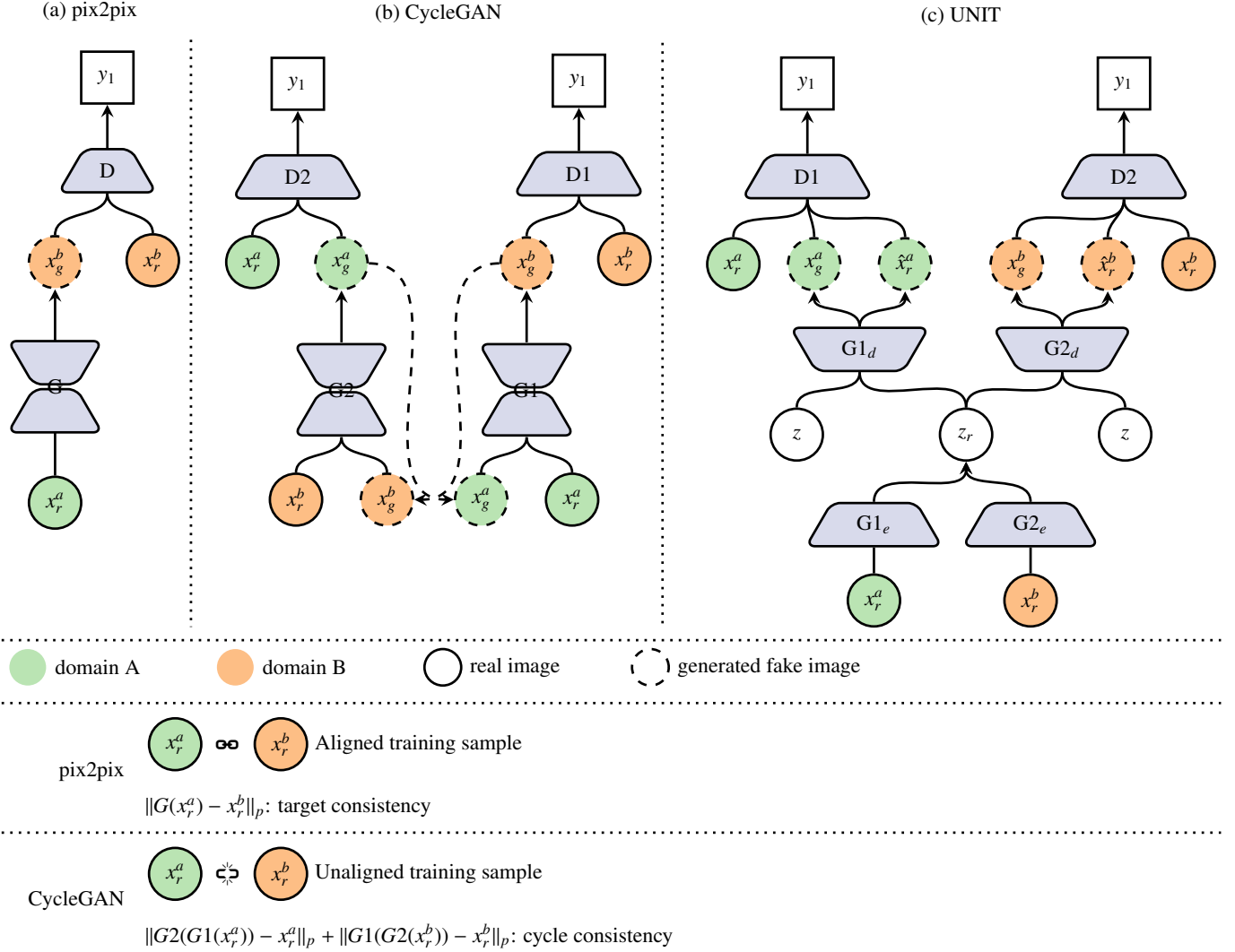
\begin{figure*}[!tb]
\centering
\resizebox{\textwidth}{!}{
\begin{tikzpicture}[every node/.style={font=\scriptsize}]
\node(pix2pix)[rectangle, label={[yshift=0.1cm](a) pix2pix}]{
\tikz{
\node{\tikz{\pic (g)  {generator}}};
\node at(g.center){G};
\node (z)[myimage, below=0.5cm of g.south, domainA]{$x_r^a$};
\node (ifake)[above=0.6cm of g.north, myimagefake,domainB]{$x^b_g$};
\draw[thick](z.north) to[out=90,in=-90] (g.south);
\draw[thick, myarrow](g.north)  to[out=90,in=-90] (ifake.south);

\node (d)[myencoder] at (0.6,2.5) {D};
\node (ireal)[below right=0.55cm and 0.3cm of d.south, myimage,domainB]{$x^b_r $};

\node (y)[above=0.5cm of d.north, myobj]{$y_1$};
\draw[thick](ifake.north) to[out=90,in=-90] (d.south);
\draw[thick](ireal.north) to[out=90,in=-90] (d.south);
\draw[thick, myarrow](d.north) to[out=90,in=-90] (y.south);


}
};

\node(CycleGAN)[rectangle, label={[yshift=0.1cm](b) CycleGAN}] at (4,0){
\tikz{
\node(g1)[inner sep=0.001pt]  {\tikz{\pic   {generator}}};
\node at(g1.center){G1};
\node (z11)[myimage, below right=0.5cm and 0.3cm of g1.south,domainA]{$x^a_r$};
\node (z12)[myimagefake, below left=0.5cm and 0.3cm of g1.south,domainA]{$x^a_g$};

\node (ifake1)[above=0.6cm of g1.north, myimagefake,domainB]{$x^b_g$};
\draw[thick](z11.north) to[out=90,in=-90] (g1.270);
\draw[thick](z12.north) to[out=90,in=-90] (g1.270);
\draw[thick, myarrow](g1.north)  to[out=90,in=-90] (ifake1.south);

\node (d1)[myencoder] at (0.6,2.5) {D1};
\node (ireal1)[below right=0.5cm and 0.3cm of d1.south, myimage,domainB]{$x^b_r $};

\node (y1)[above=0.5cm of d1.north, myobj]{$y_1$};
\draw[thick](ifake1.north) to[out=90,in=-90] (d1.south);
\draw[thick](ireal1.north) to[out=90,in=-90] (d1.south);
\draw[thick, myarrow](d1.north) to[out=90,in=-90] (y1.south);

\node(g2)[inner sep=0.001pt, left = 1cm of g1]  {\tikz{\pic  {generator}}};
\node at(g2.center){G2};
\node (z21)[myimage, below left=0.5cm and 0.3 of g2.south,domainB]{$x^b_r$};
\node (z22)[myimagefake, below right=0.5cm and 0.3 of g2.south,domainB]{$x^b_g$};

\node (ifake2)[above=0.6cm of g2.north, myimagefake,domainA]{$x^a_g$};
\draw[thick](z21.north) to[out=90,in=-90] (g2.270);
\draw[thick](z22.north) to[out=90,in=-90] (g2.270);
\draw[thick, myarrow](g2.north)  to[out=90,in=-90] (ifake2.south);

\node (d2)[myencoder, above left = 2.2cm and 0.8 cm of g2.east] {D2};
\node (ireal2)[below left=0.5cm and 0.3cm of d2.south, myimage,domainA]{$x^a_r $};

\node (y2)[above=0.5cm of d2.north, myobj]{$y_1$};
\draw[thick](ifake2.north) to[out=90,in=-90] (d2.south);
\draw[thick](ireal2.north) to[out=90,in=-90] (d2.south);
\draw[thick, myarrow](d2.north) to[out=90,in=-90] (y2.south);

\draw[thick, dashed, myarrow](ifake1.west) to[out=180,in=0] (z22.east);
\draw[thick, dashed, myarrow](ifake2.east) to[out=0,in=180] (z12.west);


}
};

\node(unit)[label={[yshift=0.1cm](c) UNIT},mybdbox]at (10,-0.6){
\tikz{
\node[mydecoder] (g1d) at (0,0) {$\text{G1}_{d}$ };
\node (z)[myimage, below left=0.5cm and 0.5cm of g1d.south]{$z$};
\node (zr)[myimage, below right=0.5cm and 1cm of g1d.south]{$z_r$};

\node (irrecon)[above right=0.5cm and 0.3cm of g1d.north, myimagefake, domainA]{$\hat{x}^a_r$};
\node (ifake)[above left=0.5cm and 0.3cm of g1d.north, myimagefake, domainA]{$x^a_g$};

\draw[thick](z.north) to[out=90,in=-90] (g1d.south);
\draw[thick](zr.north) to[out=90,in=-90] (g1d.south);

\draw[thick, myarrow](g1d.north) to[out=90,in=-90] (ifake.south);
\draw[thick, myarrow](g1d.north) to[out=90,in=-90] (irrecon.south);

\node (g1e)[myencoder, below right=1.5cm and -0.1cm of g1d.south] {$\text{G1}_{e}$};
\node (ir)[myimage, below =0.3cm of g1e.south, domainA]{$x^a_r$};
\draw[thick](ir.north)to[out=90,in=-90](g1e.south);
\draw[thick,myarrow](g1e.north)to[out=90,in=-90](zr.south);

\node (d1)[myencoder] at (-0.6,2) {D1};
\node (ireal)[left=0.3cm of ifake.west, myimage, domainA]{$x^a_r $};
\node (y)[above=0.5cm of d1.north, myobj]{$y_1$};

\draw[thick](ifake.north) to[out=90,in=-90] (d1.270);
\draw[thick](ireal.north) to[out=90,in=-90] (d1.270);
\draw[thick](irrecon.north) to[out=90,in=-90] (d1.270);

\draw[thick, myarrow](d1.north) to[out=90,in=-90] (y.south);

\node[mydecoder, right=1cm of g1d] (g2d) {$\text{G2}_{d}$ };
\node (zb)[myimage, below right=0.5cm and 0.5cm of g2d.south]{$z$};

\node (irreconb)[above right=0.5cm and 0.3cm of g2d.north, myimagefake, domainB]{$\hat{x}^b_r$};
\node (ifakeb)[above left=0.5cm and 0.3cm of g2d.north, myimagefake, domainB]{$x^b_g$};

\draw[thick](zb.north) to[out=90,in=-90] (g2d.south);
\draw[thick](zr.north) to[out=90,in=-90] (g2d.south);

\draw[thick, myarrow](g2d.north) to[out=90,in=-90] (ifakeb.south);
\draw[thick, myarrow](g2d.north) to[out=90,in=-90] (irreconb.south);

\node (g2e)[myencoder, below left=1.5cm and 0.1cm of g2d.south] {$\text{G2}_{e}$};
\node (irb)[myimage, below =0.3cm of g2e.south, domainB]{$x^b_r$};
\draw[thick](irb.north)to[out=90,in=-90](g2e.south);
\draw[thick,myarrow](g2e.north)to[out=90,in=-90](zr.south);

\node (d2)[myencoder, right=2.4cm of d1] {D2};
\node (irealb)[right=0.3cm of irreconb.east, myimage, domainB]{$x^b_r $};
\node (yb)[above=0.5cm of d2.north, myobj]{$y_1$};

\draw[thick](ifakeb.north) to[out=90,in=-90] (d2.270);
\draw[thick](irealb.north) to[out=90,in=-90] (d2.270);
\draw[thick](irreconb.north) to[out=90,in=-90] (d2.270);

\draw[thick, myarrow](d2.north) to[out=90,in=-90] (yb.south);
}
};

\draw[thick,dotted] (pix2pix.north east) -- ($(pix2pix.south east)+(0,-1)$);
\draw[thick,dotted]  (CycleGAN.north east) -- ($(CycleGAN.south east)+(0,-1)$);

\node(caption)[below=3cm of pix2pix.south west, anchor=west,mybdbox]{
\tikz{

\node(cap)[]at(0,0){
\tikz{
\node(xa)[domainA, myimage, minimum size=12pt, draw=none, label=right:{domain A}] at (0,0) {};
\node(xb)[domainB,right=55pt of xa.east, myimage, minimum size=12pt, draw=none, label=right:{domain B}] {};
\node(real)[right= 55pt of xb.east, myimage, minimum size=12pt,  label=right:{real image}]{};
\node(fake)[right= 55pt of real.east, myimage, minimum size=12pt,  label=right:{generated fake image}, dashed]{};
}
};

\node(pixcap)[anchor=west, inner sep=2pt, label=left:{pix2pix}]at(1.5,-1.2){
\tikz{
\node(capz)[ myimage,domainA]{$x^a_r $};
\node(capchain)[right=0.6cm of capz.center, rotate=45, anchor=center]{\faChain};
\node(capireal)[right=0.3cm of capchain.center, myimage,domainB, label=right:{Aligned training sample}]{$x^b_r $};
\node(loss)[below left=0.5cm and 0.2cm of capz.south west, anchor=west]{$||G(x_r^a)-x_r^b||_p$: target consistency };
}
};

\node(cyclecap)[anchor=west, inner sep=2pt, label=left:{CycleGAN}]at(1.5,-2.7){
\tikz{
\node(capz2)[myimage,domainA]{$x^a_r $};
\node(capchain2)[right=0.6cm of capz2.center, rotate=45, anchor=center]{\faChainBroken};
\node(capireal2)[right=0.3cm of capchain2.center, myimage,domainB, label=right:{Unaligned training sample}]{$x^b_r $};
\node(loss2)[below left=0.5cm and 0.2cm of capz2.south west, anchor=west]{$||G2(G1(x_r^a))-x_r^a||_p + ||G1(G2(x_r^b))-x_r^b||_p$: cycle consistency };
}
};

\draw[dotted, thick]($(cap.north west)+(0,0)$)  --++ (14, 0);
\draw[dotted, thick]($(cap.south west)+(0,-0.1)$)  --++ (14, 0);
\draw[dotted, thick]($(cap.south west)+(0,-1.6)$)   --++ (14, 0);

}
};

\end{tikzpicture}}
\caption{cGAN frameworks for image-to-image translation. pix2pix requires aligned training data whereas this constraint  is relaxed in CycleGAN  but usually suffers from performance loss. Note that in (a), we chose reconstruction loss as an example of target consistency. This supervision is task related  and can take many other different forms. (c)  It consists of two VAEGANs with shared latent vector in the VAE part. }
\label{i2i}
\end{figure*}

One of the most common conditional inputs $c$ is an  image. pix2pix, the first general purpose GAN based image-to-image translation framework was proposed by~\cite{isola2016image} (Figure~\ref{i2i} a). Further, task related supervision was introduced to the generator. For example, reconstruction loss for image restoration and Dice loss~\citep{milletari2016v} for segmentation. This form of supervision requires aligned training pairs.  \cite{zhu2017unpaired, kim2017learning} relaxed this constraint by stitching two generators together head to toe so that images can be translated between two sets of unpaired samples (Figure~\ref{i2i} b).  For the sake of simplicity, we chose CycleGAN to represent this idea in the rest of this paper. Another model named UNIT (Figure~\ref{i2i} c) can also perform unpaired image-to-image transform by combining two VAEGANs together with each one responsible for one modality but sharing the same latent space~\citep{liu2017unsupervised}.  These image-to-image translation frameworks are very popular in the medical imaging community due to their general applicability.

Other than image, the conditional input can be class labels (CGAN, Figure~\ref{ganvar} h)~\citep{mirza2014conditional}, text descriptions~\citep{zhang2017stackgan}, object locations~\citep{reed2016generative,reed2016learning}, surrounding image context~\citep{pathak2016context}, or sketches~\citep{sangkloy2016scribbler}. Note that ACGAN mentioned in the previous section also has a class conditional generator.

\subsubsection{\textbf{Varying architecture}}
Fully connected layers were used as the building block in vanilla GAN  but later on, were replaced by fully convolutional downsampling/upsampling layers in DCGAN~\citep{radford2015unsupervised}. DCGAN demonstrated better training stability hence quickly populated the literature. As shown in Figure~\ref{arch}, the generator in DCGAN architecture works on random input noise vector  by successive upsampling operations eventually generating an image from it. Two of its important  ingredients are BatchNorm~\citep{ioffe2015batch} for regulating the extracted feature scale, and LeakyRelu~\citep{maas2013rectifier} for preventing dead gradients. Very recently, \cite{miyato2018spectral} proposed a spectral normalization layer that normalized weights in the discriminator to regulate the  scale of feature response values. With the training stability   improved, some works have also incorporated residual connections into both the generator and discriminator and experimented  with much deeper networks~\citep{gulrajani2017improved, miyato2018spectral}. The work in \cite{miyato2018cgans}  proposed a projection based way to incorporate the conditional information instead of direct concatenation and found it to be beneficial in improving the  generated image's quality.

Directly generating high resolution images from a noise vector is hard, therefore some works have proposed tackling it in a progressive manner.   In LAPGAN  (Figure~\ref{ganvar} i), ~\cite{denton2015deep} proposed  a stack of  GANs, each of which adds higher frequency details into the generated image. In SGAN, a cascade of GANs is also used but each GAN generates increasingly lower level representations ~\citep{huang2017stacked}, which are compared with the hierarchical representations extracted from a discriminatively trained model. \cite{karras2017progressive} adopted an alternate way where they progressively grow the generator and discriminator by adding new layers to them rather than stacking another GAN on top of the preceding one (PGGAN). This progressive idea was also explored in conditional setting~\citep{wang2017high}.  More recently, \cite{karras2018style} proposed a style-based generator architecture (styleGAN) where instead of directly feeding the latent code $z$ to the input of the generator, they transformed this code first to an intermediate latent space and then use it to  scale and shift the normalized image feature responses computed from each convolution layer.  Similarly,~\cite{park2019semantic} proposed SPADE where the segmentation mask was injected to the generator via a spatially adaptive normalization layer. This conditional setup was found to better preserve the semantic layout of the mask than directly feeding the mask to the generator.

Schematic illustrations of the most representative   GANs are shown in Figure~\ref{ganvar}. They are GAN, CatGAN, EBGAN/BEGAN, ALI/BiGAN, InfoGAN, ACGAN,  VAEGAN, CGAN,   LAPGAN, SGAN. Three popular image-to-image translation cGANs  (pix2pix, CycleGAN, and UNIT) are shown in Figure~\ref{i2i}.   For a more in-depth review and empirical evaluation of these different variants of GAN, we refer the reader to~\citep{huang2018introduction, creswell2018generative, kurach2018gan}.

\section{Applications in Medical Imaging}\label{sect:app}
\begin{figure*}[!htb]
\centering
\begin{tikzpicture}[every node/.style={font=\scriptsize}]
\node(sagan)[label=above:{{(a) low dose CT denoising}}, inner sep=0]{
\begin{tikzpicture}
 \matrix [column sep=0.3mm, row sep=5mm,ampersand replacement=\&] {
 		 \node (p11)[inner sep=0] at (0,0){\includegraphics[width=0.165\textwidth]{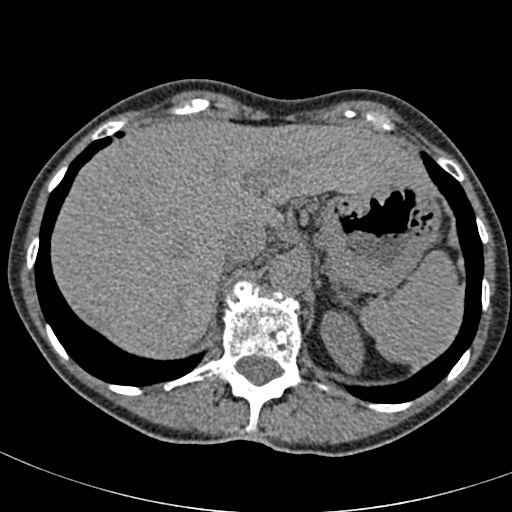}};     \&
		 \node (p12)[inner sep=0] at (0,0){\includegraphics[width=0.165\textwidth]{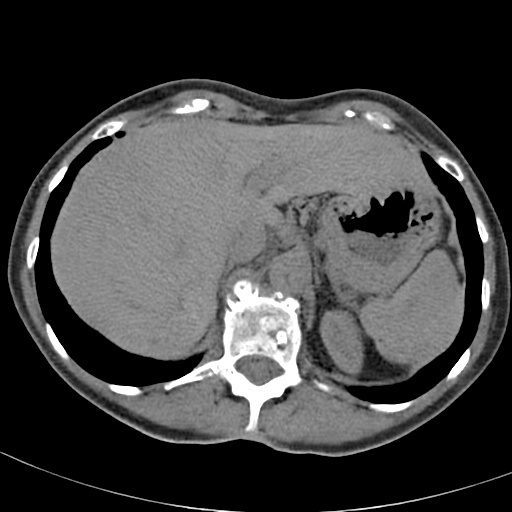}};     \&\\
                       };
  \end{tikzpicture}
};
\node(mrct)[label=above:{{(b) Cross modality transfer (MR$\rightarrow$ CT)}},right=0 of sagan.north east, anchor=north west, inner sep=0]{
\begin{tikzpicture}
 \node (p11)[inner sep=0] at (0,0){\includegraphics[width=0.33\textwidth]{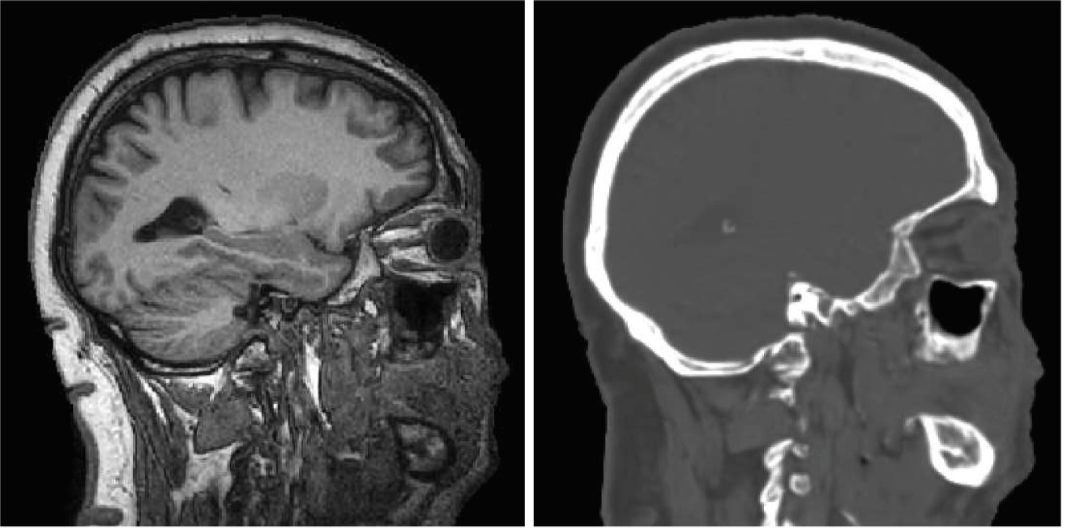}};  
  \end{tikzpicture}
};
\node(fundus)[label=above:{{(c) Vessel to fundus image}},right=0 of mrct.north east, anchor=north west,mybdboxdashed, inner sep=0]{
\begin{tikzpicture}
  \node (p11)[inner sep=0] at (0,0){\includegraphics[width=0.33\textwidth]{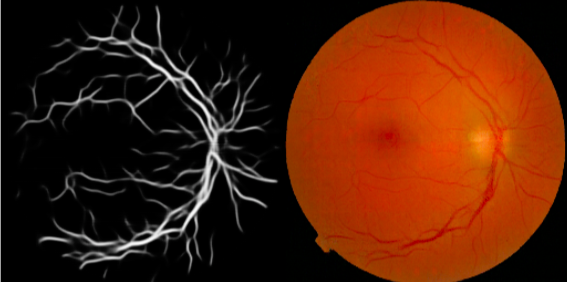}}; 
  \end{tikzpicture}
};

\node(dermo)[label=above:{(d) Skin lesion synthesis},below=0.5 cm of sagan.south west, anchor=north west,mybdboxdashed, inner sep=0]{
\begin{tikzpicture}
 \node (p11)[inner sep=0] at (0,0){\includegraphics[width=0.33\textwidth, height=0.17\textwidth]{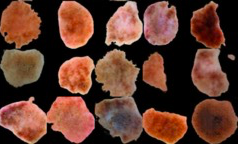}};
\end{tikzpicture}
};
\node(organ)[label=above:{{(e) Organ segmentation}},right=0 of dermo.north east, anchor=north west,mybdboxdashed, inner sep=0]{
\begin{tikzpicture}
  \node (p11)[inner sep=0] at (0,0){\includegraphics[width=0.175\textwidth]{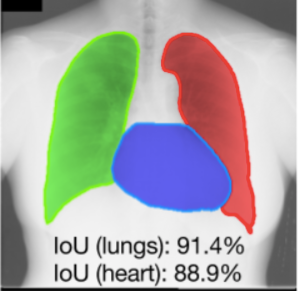}}; 
  \end{tikzpicture}
};
\node(da)[label=above:{{(f) Domain adaptation}},right=0 of organ.north east, anchor=north west,mybdboxdashed, inner sep=0]{
\begin{tikzpicture}
  \node (p11)[inner sep=0] at (0,0){\includegraphics[width=0.49\textwidth]{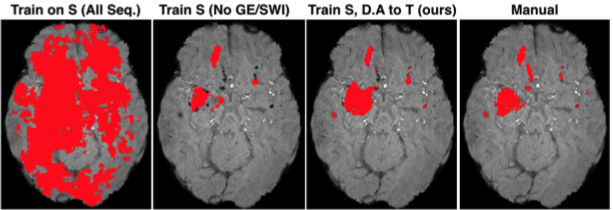}}; 
  \end{tikzpicture}
};
\node(detect)[label=above:{(g) Abnormality Detection},below=0.5 cm of dermo.south west, anchor=north west,mybdboxdashed, inner sep=0]{
\includegraphics[width=\textwidth]{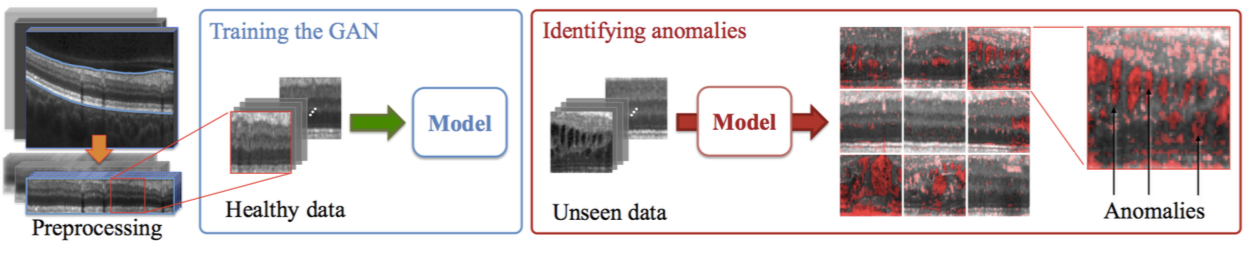}
};

\end{tikzpicture}

\caption{Example applications using GANs. Figures are directly cropped from the corresponding papers. (a) Left side shows the noise contaminated low dose CT and right side shows the denoised CT that well preserved the low contrast regions in the liver ~\citep{yi2018sharpness}. (b) Left side shows the MR image and right side shows the synthesized  corresponding CT. Bone structures were well delineated in the generated CT image ~\citep{wolterink2017deep}. (c) The generated retinal fundus image have the exact vessel structures as depicted in the left vessel map ~\citep{costa2017end}. (d) Randomly generated skin lesion from random noise (a mixture of malignant and  benign)~\citep{yi2018unsupervised}. (e) An organ (lung and heart) segmentation example on adult chest X-ray. The shapes of lung and heart are regulated by the adversarial loss~\citep{dai2017scan}.  (f) The third column shows the domain adapted brain lesion segmentation result on SWI sequence without training with the corresponding manual annotation~\citep{kamnitsas2017unsupervised}. (g) Abnormality detection of  optical coherence tomography images of the retina~\citep{schlegl2017unsupervised}. }
\label{fig:gansample}
\end{figure*}
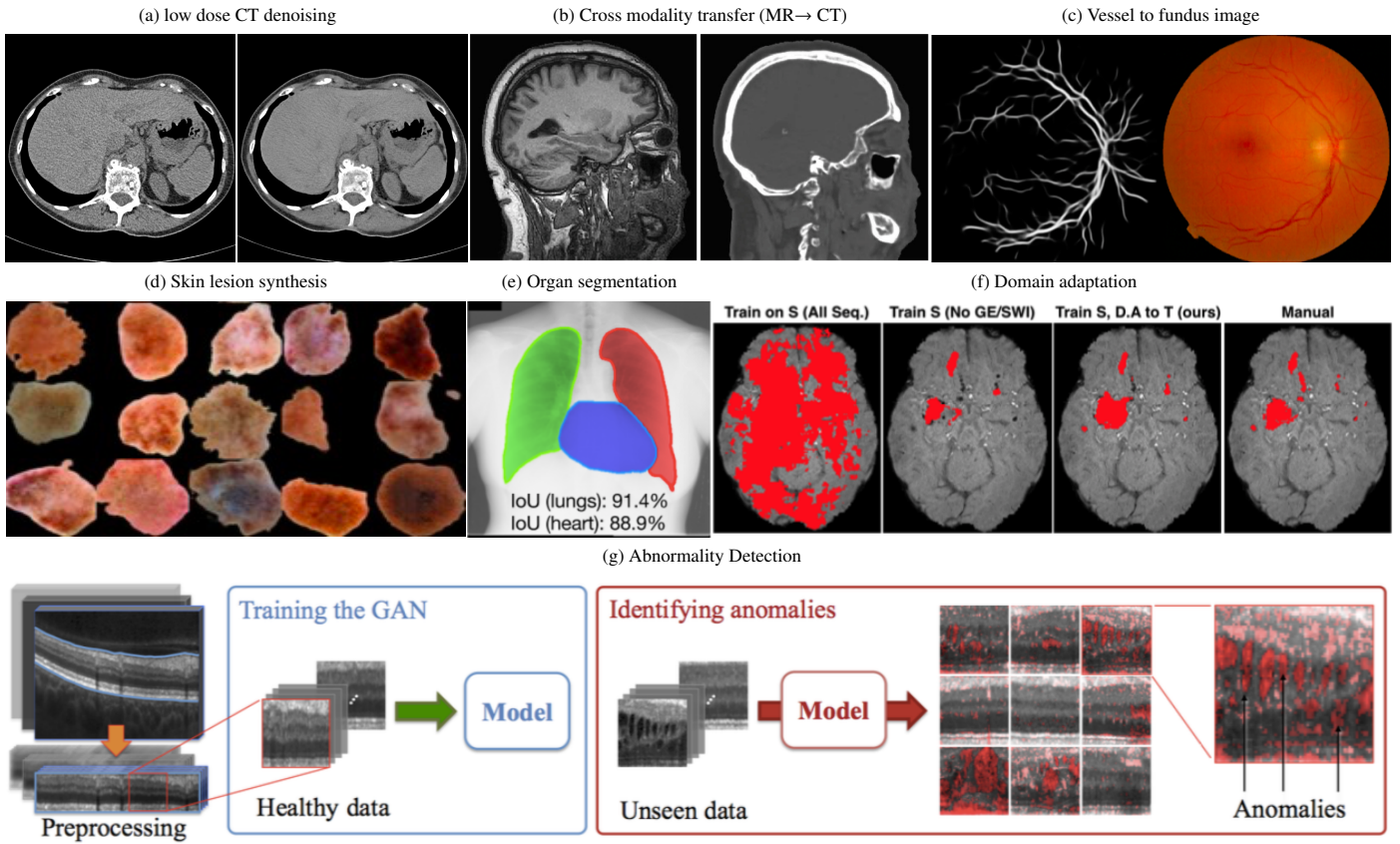

There are generally two ways  GANs are used in medical imaging. The first is focused on the generative aspect, which can help in exploring and discovering the underlying structure of training data and learning to generate new images. This property makes GANs very promising in coping with  data scarcity and patient privacy. The second focuses on the discriminative aspect, where the discriminator D can be regarded as a learned prior for normal images so that it can be used as regularizer or detector when presented with abnormal images.  Figure~\ref{fig:gansample} provides examples of  GAN related applications, with examples (a), (b), (c), (d), (e), (f) that focus on the generative aspect and example (g) that exploits the discriminative aspect. In the following subsections, in order to help the readers  find applications of their interest, we categorized all the reviewed articles into canonical tasks: reconstruction, image synthesis, segmentation, classification, detection, registration, and others.

\subsection{Reconstruction}\label{sect:recon}
Due to constraints in clinical settings, such as radiation dose and patient comfort, the diagnostic quality of acquired medical images may be limited by noise and artifacts. In the last decade, we have seen a paradigm shift in reconstruction methods changing from analytic to iterative and now to machine learning based methods. These data-driven learning based methods either learn to transfer raw sensory inputs directly to output images or serve as a post processing step for reducing image noise and removing artifacts. Most of the methods reviewed in this section are borrowed directly from the computer vision literature that formulate post-processing as an image-to-image translation problem where the conditioned inputs of cGANs are compromised in certain forms, such as low spatial resolution, noise contamination, under-sampling, or aliasing. One exception is for MR images where the Fourier transform is used to incorporate the raw K-space data into the reconstruction.

The basic pix2pix framework has been used for low dose CT denoising~\citep{wolterink2017generative}, MR reconstruction~\citep{chen2018efficient, kim2018improving, dar2018synergistic, shitrit2017accelerated}, and PET denoising~\citep{wang20183d}.   A pretrained VGG-net~\citep{simonyan2014very} was further incorporated into the optimization framework to ensure perceptual similarity~\citep{yang2017low, yu2017deep, yang2018dagan, armanious2018medgan, mahapatra2017retinal}. \cite{yi2018sharpness} introduced a pretrained sharpness detection network   to explicitly constrain the sharpness of the denoised CT  especially for low contrast regions.  \cite{mahapatra2017retinal} computed a  local saliency map  to highlight blood vessels in superresolution process of retinal fundus imaging. A similar idea was explored by~\cite{liao2018adversarial} in sparse view CT reconstruction. They compute a focus map to modulate the reconstructed output to ensure that the network focused on important regions. Besides ensuring image domain data fidelity, frequency domain data fidelity is also imposed when raw K-space data is available in MR reconstruction~\citep{quan2018compressed, mardani2017deep, yang2018dagan}. 

Losses of other kinds have been used to highlight local image structures in the reconstruction, such as  the saliency loss  to reweight each pixel's importance based on its perceptual relevance~\citep{mahapatra2017retinal} and the style-content loss in PET denoising~\citep{armanious2018medgan}. In image reconstruction of moving organs, paired training samples are hard to obtain. Therefore, \cite{ravi2018adversarial} proposed a physical acquisition based loss to regulate the generated image structure for endomicroscopy super resolution and \cite{kang2018cycle} proposed to use CycleGAN together with an identity loss in the denoising of cardiac CT. \cite{wolterink2017generative} found that in low dose CT denoising, meaningful results can still be achieved when removing the image domain fidelity loss from the pix2pix framework,  but the local image structure can be altered. Papers relating to medical image reconstruction are summarized in Table~\ref{table:recon}. 


\renewcommand\arraystretch{0.8}
\begin{table*}[!tb]
\tiny
\caption{Medical image reconstruction publications. In the second column, * following the method denotes some modifications on the basic framework either on the network architecture or on the employed losses. A brief description of the losses, quantitative measures and datasets can be found in Table~\ref{table:loss},~\ref{table:metric} and~\ref{table:dataset}.  In the last column, symbol \yes~and \no~denotes whether the corresponding literature used paired training data or not. All studies were performed in 2D unless otherwise mentioned.}
\centering

\begin{tabular*}{\textwidth}{@{}llllll@{}}
		\toprule
    		\textbf{Publications}	& \textbf{Method} & \textbf{Losses} &\textbf{Dataset}	&\textbf{Quantitative Measure} &\textbf{Remarks}\\\midrule
		\textbf{\textit{CT}}&&&\\
		\midrule
		\cite{wolterink2017generative} 		& pix2pix* 		&L\ladv, \limage  			& --					&M\tss 							&[\no] [3D] Denoising  \\
		\cite{yi2018sharpness} 			& pix2pix*			&L\ladv, \limage, \lsharp	 	&D\piglet				&M\psnr, \ssim, \lineprofile, \noiselevel	&[\yes]  Denoising     \\
		\cite{yang2017low}  				& pix2pix*			&L\ladv, \limage, \lpercep 		&D\ldct 				& M\psnr, \ssim, \noiselevel  			&[\yes] [3D] [Abdomen] Denoising    \\
		\cite{kang2018cycle}   			& CycleGAN*		&L\ladv, \lcycle, \lreg			& --					& M\psnr, \noiselevel 				&[\no] [Coronary] Denoising  CT\\
		 \cite{you2018structure}			& pix2pix*			&L\ladv, \limage, \lstructure 	 & D\ldct 				& M\humanobserver, \mae, \psnr, \ssim, \noiselevel   &[\yes]  [3D]  Denoising  \\ 	
		 \cite{tang2018ct}  & SGAN			&L\ladv, \limage, \lpercep 	 	& -- 					& M\segmentation							 &[\yes] Denoising, contrast enhance   \\ 	
		 \cite{shan20183}  			& pix2pix*			&L\ladv, \lpercep 	 		& D\ldct  				& M\perceptual, \textureloss, \psnr, \ssim			&[\no] [3D] Denoising transfer from 2D  \\ 	
		 \cite{liu2019tomogan}  		& pix2pix*			&L\ladv, \limage, \lpercep 	 	& --  					& M\ssim, \lineprofile					&[\yes] Denoising, using adjacent slice  \\ 	
		\cite{liao2018adversarial}  		& pix2pix*			&L\ladv, \limage, \lpercep		& --					& M\mae, \psnr, \ssim				&[\yes] Sparse view CT reconstruction\\	
		\cite{wang2018conditional}  	& pix2pix			&L\ladv, \limage			& --					& M\segmentation					&[\yes] Metal artefact reduction  cochlear implants\\
		\cite{you2018ct}			& CycleGAN*		&L\ladv, \limage, \lselfreg			& D\ldct					& M\psnr, \ssim, \ifc					&[\yes] Superresolution, denoising\\	
		\cite{gans2018sparse}		& pix2pix*			&L\ladv , \limage 				& --					& M\mae, \psnr 					&[\yes] Sparse view CT reconstruction\\	
		  \cite{armanious2018adversarial}	& pix2pix*			&  L\ladv, \limage,   \lpercep, \lstylecontent			& --					&M\mae, \psnr, \ssim, \uqi,  	&[\yes] Inpainting\\		
		\midrule	
		\textbf{\textit{MR}} &&&\\	
		\midrule	
		  \cite{quan2018compressed}		& pix2pix*			& L\ladv, \limage, \lfreq 			&D\ixi, \dsbonefive, \mridata&  M\mae, \psnr, \ssim		&[\yes] Under-sampled K-space\\
		  \cite{mardani2017deep}			& pix2pix*			&  L\ladv, \limage, \lfreq  			& --					&M\humanobserver			&[\yes] Under-sampled K-space\\
		 \cite{yu2017deep}				& pix2pix*			&  L\ladv, \limage, \lpercep		& D\ixi, \miccaigcthree 	& M\mae, \psnr, \ssim, \lineprofile	&[\yes] Under-sampled K-space\\
		 \cite{yang2018dagan}			& pix2pix*			& L\ladv, \limage, \lpercep, \lfreq  	& D\miccaigcthree, \brainles  & M\mae, \psnr, \ssim, \lineprofile&[\yes]  Under-sampled K-space\\
		 \cite{sanchez2018brain} 			& pix2pix*			& L\ladv, \limage, \lgrad 			&D\adni				& M\psnr, \ssim 			&[\yes] [3D] Superresolution\\
		 \cite{chen2018efficient} 			& pix2pix*			&  L\ladv, \limage 				& --					&M\mae, \psnr, \ssim 		&[\yes] [3D] Superresolution\\
		 \cite{kim2018improving} 			& pix2pix*			&  L\ladv, \limage 				& D\bratsfive			&M\humanobserver, \mae, \ssim, \cbr &[\yes] Superresolution\\
		 \cite{dar2018synergistic} 			& pix2pix*			&  L\ladv, \limage				&D\ixi, \bratsfive, \midas 	& M\psnr, \ssim  			&[\yes] Under-sampled K-space\\
		  \cite{shitrit2017accelerated}		& pix2pix*			&  L\ladv, \limage 				& --					&M\psnr  					&[\yes] Under-sampled K-space\\
		  \cite{ran2018denoising}			& pix2pix*			&  L\ladv, \limage, \lpercep 		& D\ixi				&M\psnr ,\ssim 				&[\yes] [3D] Denoising\\
		  \cite{seitzer2018adversarial}		& pix2pix*			&  L\ladv, \limage, \lpercep 		& --					&M\humanobserver, \psnr, \sis 	&[\yes] Two stage\\
		  \cite{abramian2018refacing}		& CycleGAN		&  L\ladv, \limage, \lcycle			& D\ixi					&M\ssim, \cc 	&[\yes] Facial anonymization problem\\
		  \cite{armanious2018adversarial}	& pix2pix*			&  L\ladv, \limage, \lpercep, \lstylecontent			& --					&M\mae, \psnr,  \ssim, \uqi 	&[\yes] Inpainting\\
		  \cite{oksuz2018cardiac}			& pix2pix*			&  L\ladv, \limage				& D\biobank					&M\mae, \psnr, \ssim   	&[\yes] Motion correction\\
		  \cite{zhang2018multi}			& pix2pix*			&  L\ladv, \limage, \lpercep, \lselfreg				& --					&M\psnr, \ssim  	&[\yes] Directly in complex-valued k-space data\\
		  \cite{armanious2018retrospective}			& pix2pix*			&  L\ladv, \limage,  \lpercep, \lstylecontent			& --					&M\ssim, \vif, \uqi, \lpips 	&[\yes] Motion correction\\

		  \midrule	
		 \textbf{\textit{PET}}&&&\\	
		 \midrule
		\cite{wang20183d}  				& cascade cGAN	& L\ladv, \limage 					& --					&M\mae, \psnr, \suv   		&[\yes] [3D]\\
		 \cite{armanious2018medgan} 		& pix2pix*			&L\ladv, \limage, \lpercep, \lstylecontent	 & --					&M\humanobserver, \mae, \psnr, \ssim, \vif, \uqi, \lpips&[\yes] \\\midrule	
		 \textbf{\textit{Retinal fundus imaging }}&&&\\		
		 \midrule
		  \cite{mahapatra2017retinal} 	& pix2pix*			& L\ladv, \limage, \lpercep, \lsaliency 	& --					&M\mae, \psnr, \ssim		&[\yes] Superresolution	\\\midrule	
		 \textbf{\textit{Endomicroscopy} }&&&\\
		 \midrule
		 \cite{ravi2018adversarial}		& pix2pix*			& L\ladv, \lphysical, \lreg  				& --					&M\gcf, \ssim			&[\no]  Superresolution \\
		 \bottomrule		
\end{tabular*}
\label{table:recon}
\end{table*}


It can be noticed that the underlying methods are almost the same for all the reconstruction tasks. MR is special case as it has a well defined forward and backward operation, i.e. Fourier transform, so that raw K-space data can be incorporated. The same methodology can potentially be applied to incorporate the sinogram data in the CT reconstruction process but we have not seen any research using this idea as yet probably because the sinogram data is hard to access. The more data used, either raw K-space or image from other sequence, the better are the reconstructed results. In general, using adversarial loss produces more visually appealing results than using pixel-wise reconstruction loss alone. But using adversarial loss to match the generated and real data distribution may make the model hallucinate unseen structures. Pixel-wise reconstruction loss  helps to combat this problem if paired samples are available, and if the model was trained on all healthy images but employed to reconstruct images with pathologies, the hallucination problem will still exist due to domain mismatch. \cite{cohen2018distribution} have conducted extensive  experiments  to investigate this problem and suggest that reconstructed images should not be used for direct diagnosis by radiologists unless the model has been properly verified.  

However, even though the dataset is carefully curated to match the training and testing distribution, there are other problems in further boosting performance. We have seen various different losses introduced to the pix2pix framework as shown in Table~\ref{table:loss} to improve the reconstructed fidelity of local structures. There is, however, no reliable way of comparing  their effectiveness except for relying on human observer or downstream image analysis tasks. Large scale statistical analysis by human observer is currently lacking for  GAN based  reconstruction methods.  Furthermore, public datasets used for image reconstruction are not tailored towards further medical image analysis, which leaves a gap between  upstream reconstruction and downstream analysis tasks. New reference standard datasets should be created for better comparison of these GAN-based methods.


\renewcommand\arraystretch{0.8}
\renewcommand{\baselinestretch}{1.3}

\begin{table*}[!tb]
\centering
\tiny
\caption{A brief summary of different losses used in the reviewed publications in Tables~\ref{table:recon} and~\ref{table:cross}. The  third column specifies conditions to be fulfilled in order to use the corresponding loss. L in the first column stands for loss. }
\begin{tabular*}{\textwidth}{@{\extracolsep{\fill} }lllp{0.6\textwidth}}
		\toprule
		Abbr. &\textbf{Losses}	& \textbf{Requirement}  &\textbf{Remarks} \\\midrule
		L\ladv		&$\ML_{\text{adversarial}}$ 			&--						& Adversarial loss introduced by the discriminator, can take the form of cross entropy loss, hinge loss, least square loss etc. as discussed in Section~\ref{sec:objd} \\
		 L\limage		&$\ML_{\text{image}}$		 		&Aligned training pair		& Element-wise data fidelity loss in image domain to ensure structure  similarity to the target  when aligned training pair is provided  \\
		 L\lcycle		&$\ML_{\text{cycle}}$ 				&--						& Element-wise loss  to ensure self-similarity during cycled transformation when unaligned training pair is provided \\
		 L\lgrad		&$\ML_{\text{gradient}}$ 				&Aligned training pair		& Element-wise loss in the gradient domain to emphasize edges\\
		 L\ledge		&$\ML_{\text{edge}}$ 				&Aligned training pair		& Similar to $\ML_{\text{gradient}}$ but using gradient feature map as a weight to image pixels \\
		L\lsharp		&$\ML_{\text{sharp}}$ 				&Aligned training pair		& Element-wise loss in a feature domain computed from a pre-trained network, which is expected to be the image sharpness  with focus on low contrast regions  \\
		L\lshape		&$\ML_{\text{shape}}, \ML_{\text{seg}}$    &Annotated pixel-wise label	& Loss  introduced by a segmentor to ensure faithful reconstruction of anatomic regions \\
		L\lpercep		&$\ML_{\text{perceptual}}$ 			&Aligned training pair		& Element-wise loss in a feature domain computed from a pre-trained network which expected to conform to visual perception \\
		L\lstructure	&$\ML_{\text{structure}}$ 				&Aligned training pair		& Patch-wise loss in the image domain computed with SSIM which claims to better conform to human visual system  \\
		L\lstructuretwo	&$\ML_{\text{structure2}}$ 			&Aligned  pair				& MIND~\citep{heinrich2012mind} as used in image registration for two images with the same content from different modality  \\
		L\lstylecontent  &$\ML_{\text{style-content}}$ 			&Aligned training pair		& Style and content loss to ensure similarity of image style and content. Style is defined as the Gram matrix which is basically the correlation of low-level features \\
		 L\lselfreg		&$\ML_{\text{self-reg}}$ 				&--						& Element-wise loss in image domain to ensure structure  similarity to the input. Useful in denoising since the two have similar underlying structure \\
		L\lsteer		&$\ML_{\text{steer}}$ 				&Aligned training pair		&  Element-wise loss in a feature domain which is computed from steerable filters with focus on vessel-like structures \\
		L\lclassify		&$\ML_{\text{classify}}$ 				&Aligned image-wise label 	& Loss introduced by a classifier to get semantic information \\
		L\lfreq		&$\ML_{\text{frequency}}$ 			&Aligned training pair		& Element-wise loss in frequency domain (K-space) used in MR image reconstruction \\
		L\lkl			&$\ML_{\text{KL}}$ 					&--						& Kullback--Leibler divergence which is commonly seen in variational inference to ensure closer approximation to the posterior distribution \\
		L\lsaliency		&$\ML_{\text{saliency}}$ 				&Aligned training pair		&Element-wise loss in a feature domain which is expected to be the saliency map\\
		L\lphysical		&$\ML_{\text{physical}}$ 				&Physical model			& Loss introduced by a physical image acquisition model\\
		 L\lreg		&$\ML_{\text{regulation}}$ 			&--						& Regulate the generated image contrast by keeping the mean value across row and column unchanged   \\
		 \bottomrule			
\end{tabular*}
\label{table:loss}
\end{table*}

\renewcommand\arraystretch{0.8}
\renewcommand{\baselinestretch}{1.2}
\begin{table*}[!tb]
\centering
\tiny
\caption{A brief summary of quantitative measures used in the reviewed publications listed  in Tables~\ref{table:recon},~\ref{table:uncond},~\ref{table:cross} and~\ref{table:othersyn}. }
\resizebox{\textwidth}{!}{
\begin{tabular*}{\textwidth}{@{}llp{0.5\textwidth}}
		\toprule
		Abbr. &\textbf{Measures} &\textbf{Remarks} \\\midrule
		\multicolumn{3} {@{}l} {\textbf{\textit{Overall image quality without reference}} }\\
		\midrule
		M\humanobserver										& Human observer 								 	& Gold standard but costly and hard to scale  \\
		M\kerneldensity~~\cite{breuleux2011quickly} 					& Kernel density function									&Estimate the probability density of the generated data and compute the log likelihood of real test data under this distribution  \\
		M\inceptionscore~~\cite{salimans2016improved} 				& Inception score 									& Measure the generated images' diversity and visual similarity to the real images with the pretrained Inception model  \\
		M\js~~\cite{goodfellow2014generative}						& JS divergence  									&Distance measure between two distributions (used for comparison between normalized color histogram computed from a large batch of image samples)  \\
		M\wasdiscolorhist										& Wasserstein distance 								&Distance measure between two distributions (used for comparison between normalized color histogram computed from a large batch of image samples)  \\
		M\gcf~\cite{matkovic2005global}						& GCF 								&Global contrast factor\\
		M\qv~\cite{kohler2013automatic}						& $Q_v$					&Vessel-based quality metric (noise and blur) for fundus image   \\
		M\isc~\cite{niemeijer2006image}						& ISC					& Image structure clustering, a trained classifier based  to differentiate normal from low quality fundus images  \\
		M\perceptual~\cite{shan20183}							& Perceptual loss				&Difference of features extracted from a pre-trained VGG net  \\
		M\textureloss~\cite{shan20183}							& Texture loss					& Gram matrix which is basically the correlation of low-level features, defined as style in style transfer literature   \\
		\midrule
		\multicolumn{3} {@{}l} {\textbf{\textit{Overall image quality with respect to a groundtruth}} }\\
		\midrule
		M\mae												& NMSE/MAE/MSE									&(Normalized) mean absolute/square error with respect to a given groundtruth \\
		M\psnr												& PSNR/SNR 										&(Peak) signal to noise ratio with respect to a given groundtruth \\
		M\ssim~\cite{wang2004image} 							& SSIM											& Structural similarity with respect to a given groundtruth \\
		M\vif~\cite{sheikh2004image} 								&VIF						&Visual information fidelity  with regard to a given groundtruth \\
		M\uqi~\cite{wang2002universal} 							& UQI 						&Universal quality index with regard to a given groundtruth \\
		M\ifc~\cite{sheikh2005information} 							& IFC 						&Information Fidelity Criterion \\
		M\fsim~\cite{zhang2011fsim}								& FSIM						&A low-level feature based image quality assessment  metric with regard to a given groundtruth \\
		M\lpips~\cite{zhang2018unreasonable}						& LPIPS   				&Learned perceptual image patch similarity \\
		M\mi~\cite{pluim2003mutual}								&Mutual information  								&Commonly used in cross modality registration in evaluating the alignment of two images  \\
		M\medianintensity			&  NMI/MI											&(Normalized) median intensity, used to measure color consistancy of histology images  \\
		M\cc~\cite{lee1988thirteen}	&  Cross correlation									&Global correlation between two images  \\
		M\clinical~\cite{low2010gamma}					&  Clinical measure				&Dose difference, gamma analysis for CT  \\
		M\sis~\cite{seitzer2018adversarial}					&  SIS						&Semantic interpretability score, essentially the dice loss of a pre-trained downstream segmentor  \\
		\midrule
		\multicolumn{3} {@{}l} {\textbf{\textit{Local image quality}} }\\
		\midrule
		M\lineprofile				& Line profile 										&Measure the loss of spatial resolution \\
		M\noiselevel				& Noise level										&Standard deviation of intensities in a local smooth region \\
		M\cbr					& CBR											&Contrast to background ratio, measure the local contrast loss\\
		M\suv~\cite{kinahan2010positron}					& SUV						& Standard uptake value, a clinical measure in oncology for local interest region, should not vary too much in reconstruction\\
		M\nps										& NPS						& Noise power spectrum\\
		\midrule
		\multicolumn{3} {@{}l} {\textbf{\textit{Image quality analysis by auxiliary task} }}\\
		\midrule
		M\tss					& Task specific statistics 								&Down stream task (e.g. for coronary calcium quantification)  \\
		M\classification				& Classification 									&Down stream task  \\
		M\lesiondetection			& Detection 										&Down stream task (e.g. for lesion/hemorrhage)   \\
		M\segmentation			& Segmentation 									&Down stream task  \\
		M\cmr					& Cross modality registration							&Down stream task  \\
		M\depth					& Depth estimation									&Down stream task  \\
		\bottomrule		
\end{tabular*}
}
\label{table:metric}
\end{table*}

\subsection{Medical Image Synthesis}\label{sect:mis}
Depending on institutional protocols, patient consent may be  required if diagnostic images are intended to be used in a publication or released into the public domain~\citep{dysmorphology2000informed}. GANs are widely for medical image synthesis.  This helps overcome the privacy issues related to diagnostic medical image data and tackle the insufficient number of positive cases of each pathology.  Lack of experts annotating medical images poses another challenge for the adoption of supervised training methods.  Although there are ongoing collaborative efforts  across multiple healthcare agencies  aiming  to build large open access datasets, e.g. Biobank, the National Biomedical Imaging Archive (NBIA), The Cancer Imaging Archive (TCIA) and Radiologist Society of North America (RSNA), this issue remains and constrains  the number of images researchers might have access to.

Traditional ways to augment training sample include scaling, rotation, flipping, translation, and elastic deformation~\citep{simard2003best}. However, these transformations do not account for variations resulting from different imaging protocols or sequences, not to mention variations in the size, shape, location and appearance of specific pathology. GANs provide a more generic solution and have been used in numerous works for augmenting training images with promising results.

\subsubsection{Unconditional Synthesis}\label{sect:us}
Unconditional synthesis refers to image generation from random noise without any other conditional information. Techniques commonly adopted in the medical imaging community include DCGAN, WGAN, and PGGAN due to their good training stability. The first two methods can handle an image resolution of up to $256\times256$ but if higher resolution images are desired, the progressive technique proposed in PGGAN is a choice.  Realistic images can be  generated by directly using the author released code base  as long as the  variations between images are not too large, for example, lung nodules and liver lesions. To make the generated images useful for downstream tasks, most studies trained a separate generator  for each individual class; for example,~\cite{frid2018gan} used three DCGANs  to generate synthetic samples for three classes of liver lesions (cysts, metastases, and hemangiomas); generated samples were found to be beneficial to the lesion classification task with both improved  sensitivity and specificity when combined with real training data. 
~\cite{bermudez2018learning}  claimed that neuroradiologists found generated MR images to be of comparable quality to real ones, however, there were discrepancies in anatomic accuracy. Papers related to unconditional medical image synthesis are summarized  in Table~\ref{table:uncond}.

\renewcommand\arraystretch{0.8}

\begin{table*}[htp]
\centering
\tiny
\caption{Unconditional medical image synthesis publications. A brief description of the quantitative measures and datasets can be found in Tables~\ref{table:metric} and~\ref{table:dataset}.}
\resizebox{\textwidth}{!}{
\begin{tabular}{@{}llllp{0.5\textwidth}}
	\toprule
		\textbf{Publications}	&	\textbf{Method}	&\textbf{Dataset}  & \textbf{Measures}  	& \textbf{Remarks}\\\midrule
		 \textbf{\textit{CT}}				&	&&&\\
		 \midrule
		 \cite{chuquicusma2017fool}		& DCGAN 		&D\lidc		&M\humanobserver							&[Lung nodule]\\
		 \cite{frid2018gan} 				& DCGAN	/ACGAN	&--			&M\classification							&[Liver lesion] Generating each lesion class separately (with DCGAN) is than generating all classes at once (using ACGAN)\\
		\cite{bowles2018gan}			& PGGAN 		&--			&M\segmentation							&[Brain] Joint learning of image and segmentation map  \\
		\midrule
		\textbf{\textit{MR}}				&&&&\\
		\midrule
		\cite{calimeri2017biomedical}  		& LAPGAN 		& --			&M\humanobserver, \kerneldensity, \inceptionscore	& [Brain]	\\
		 \cite{zhang2017semi}			&Semi-Coupled-GAN&--			&M\classification							&[Heart] Two generators coupled with a single  discriminator which
outputted both a distribution over the image data source and class labels\\
		 \cite{han2018gan} 				& WGAN 			&D\bratssix	&M\humanobserver							&[Brain] \\
		\cite{beers2018high}  			& PGGAN 		&D\bratsseven	&-										&[Brain] \\
		\cite{bermudez2018learning}		& DCGAN 		&D\blsa		&M\humanobserver							&[Brain] \\
		\cite{mondal2018few}			& DCGAN* 		&D\bratsthree, \iseg		&M\segmentation				&[Brain] Semi-supervised training with labeled, unlabeled, generated data  \\
		\cite{bowles2018gan}			& PGGAN 		&--			&M\segmentation							&[Brain] Joint learning of image and segmentation map  \\
		\midrule
		 \textbf{\textit{X-ray}}			&	&&&\\
		 \midrule
		 \cite{salehinejad2017generalization}&DCGAN			&--			&M\classification							&[Chest] Five different GANs to generate five different classes of chest disease\\
		 \cite{madani2018semi}			&DCGAN			&D\plco		&M\classification							&[Chest] Semi-supervised DCGAN can achieve performance comparable with a traditional supervised CNN with an order of magnitude less labeled data \\
		 \cite{madani2018chest} 			& DCGAN 		&D\plco		&M\classification							& [Chest] Two GANs to generate normal and abnormal chest X-rays separately\\\midrule
		 \textbf{\textit{Mammography}}			&&&&\\
		 \midrule
		 \cite{korkinof2018high} 			& PGGAN & -- 		& --			&--\\\midrule
		 \textbf{\textit{Histopothology}}			&&&&\\
		 \midrule
		 \cite{hu2017unsupervised} 		& WGAN+infoGAN	&D\celldetect 			&M\classification, M\segmentation			& Cell level representation learning\\\midrule
		\textbf{\textit{Retinal fundus imaging}	}		&	&&&\\
		\midrule
		\cite{beers2018high} 			& PGGAN 		& --			&--										&--\\
		\cite{lahiri2017generative}			& DCGAN 		&D\drive		&M\classification							& Semi-supervised DCGAN can achieve performance comparable with a traditional supervised CNN with an order of magnitude less labeled data\\
		\cite{lahiri2018retinal}			& DCGAN 		&D\drive, \stare		&M\classification							& Extend the above work by adding an unsupervised loss into the discriminator\\
		\midrule
		\textbf{\textit{Dermoscopy}}				& 	&&&\\
		\midrule
		\cite{baur2018melanogans}		& LAPGAN 		&D\isicseven 	&M\js, \mae								&--\\	
		\cite{baur2018generating}			& PGGAN 		&D\isiceight 	&M\humanobserver							&--\\	
		 \cite{yi2018unsupervised} 		& CatGAN$+$ WGAN &D\isicsix, \phtwo&M\classification							& Semi-supervised skin lesion feature representation learning\\\bottomrule		
\end{tabular}
}
\label{table:uncond}
\end{table*}

\subsubsection{Cross modality synthesis}\label{crossmodsyn}
Cross modality synthesis (such as generating CT-like images based on MR images) is deemed to be useful for multiple reasons, one of which  is to reduce the  extra acquisition time and cost. Another reason  is to generate new training samples with  the appearance being constrained by the anatomical structures delineated  in the available modality. Most of the methods reviewed in this section share many similarities to those in Section~\ref{sect:recon}. pix2pix-based frameworks are used in cases where different image modality data can be co-registered to ensure data fidelity. CycleGAN-based frameworks are used to handle more general cases where registration is challenging such as in cardiac applications. In a study by \cite{wolterink2017deep} for brain CT image synthesis from MR image, the authors found that training using unpaired images was even better than using aligned images. This  most likely resulted from the fact that rigid registration could not very well handle local alignment in the throat, mouth, vertebrae, and nasal cavities.  ~\cite{hiasa2018cross} further incorporated gradient consistency loss in the training to improve accuracy at the boundaries. ~\cite{zhang2018translating} found that using only cycle loss in the cross modality synthesis was insufficient to mitigate geometric distortions in the transformation. Therefore, they employed a shape consistency loss that is obtained from two segmentors (segmentation network).  Each segmentor segments the corresponding image modality into semantic labels and   provides implicit shape constraints on the anatomy during translation. To make the whole system end-to-end trainable, semantic labels of training images from both modalities are required. ~\cite{zhang2018task} and~\cite{chen2018semantic}  proposed using a segmentor also in the cycle transfer using  labels in only one modality. Therefore, the segmentor is trained offline and fixed during the training of the image transfer network. As reviewed in Section~\ref{sect:bg}, UNIT and CycleGAN are two equally valid frameworks for unpaired cross modality synthesis. It was found that these two frameworks  performed almost equally well for the transformation between T1 and T2-weighted MR images~\citep{welander2018generative}. Papers related to cross modality medical image synthesis are summarized in Table~\ref{table:cross}.

\renewcommand\arraystretch{0.8}
\begin{table*}[!tp]
\centering
\tiny
\caption{Cross modality image synthesis publications. In the second column, * following the method denotes some modifications on the basic framework either on the network architecture or on the employed losses. A brief description of the losses, quantitative evaluation measures and datasets can be found in Tables~\ref{table:loss},~\ref{table:metric} and~\ref{table:dataset}.  In the last column, symbol \yes~and \no~denotes whether the corresponding literature used paired training data or not. }

\resizebox{0.8\textwidth}{!}{
\begin{tabular}{@{}lllllll@{}l}
	\toprule
		 \textbf{Pulications}	& \textbf{Method} & \textbf{Loss} &\textbf{Dataset} & \textbf{Measures}  &\textbf{Remarks}\\\midrule
		 \textbf{\textit{MR $\rightarrow$ CT }}				&&&&\\
		 \midrule
        		\cite{nie2017medical, nie2018medical}									& Cascade GAN 				&L\ladv, \limage, \lgrad				&D\adni			&M\mae, \psnr 					&[\yes]Brain; Pelvis  \\
		\cite{emami2018generating} 	           									&  cGAN 						&L\ladv, \limage  					& --				&M\mae, \psnr, \ssim 			&[\yes]Brain 	 \\\midrule
		 \textbf{\textit{CT  $\rightarrow$ MR 	}}			&&&&\\
		 \midrule
			 	\cite{jin2018deep}											& CycleGAN					&L\ladv, \limage, \lcycle				& --				&M\mae, \psnr 					&[\no]  Brain \\
			 	\cite{jiang2018tumor}										& CycleGAN*					&L\ladv, \limage, \lcycle, \lshape, \lpercep	& D\nsclc			&M\segmentation 				&[\no]  Lung \\
				\midrule
		 \textbf{\textit{MR $\leftrightarrow$ CT }}&&&&\\
		 \midrule
		\cite{chartsias2017adversarial} 											&CycleGAN	 				&L\ladv, \lcycle						&D\wholeheart		&M\segmentation				&[\no	] Heart\\
		\cite{zhang2018translating} 											&CycleGAN*					&L\ladv, \lcycle, \lshape				& --				&M\segmentation 				&[\no][3D] Heart \\
		\cite{huo2017adversarial} 												&CycleGAN*					&L\ladv, \lcycle, \lshape				&--				&M\segmentation 				&[\no] Spleen \\
		\cite{chartsias2017adversarial}						 					&CycleGAN					&L\ladv, \lcycle						&--				&M\segmentation 				&[\no]  Heart \\
		\cite{hiasa2018cross} 							 					&CycleGAN*					&L\ladv, \lcycle, \lgrad				&--				&M\mi,  \segmentation  &[\no] Musculoskeletal\\
		\cite{wolterink2017deep}						 						&CycleGAN					&L\ladv, \lcycle						&-- 				&M\mae, \psnr  				&[\no]  Brain\\
		\cite{huo2018synseg}						 						&CycleGAN					&L\ladv, \lcycle, \lshape				&-- 				&M\segmentation 				&[\no]  Abdomen\\
		\cite{yang2018unpaired}						 						&CycleGAN*					&L\ladv, \limage, \lcycle, \lstructuretwo	&-- 				&M\mae, \psnr, \ssim				&[\no]  Brain\\	
		\cite{maspero2018dose}						 						&pix2pix						&L\ladv, \limage					&-- 				&M\mae, \clinical				&[\yes]  Pelvis\\			
		\midrule
		 \textbf{\textit{CT  $\rightarrow$ PET 	}}			&&&&\\
		 \midrule
		\cite{bi2017synthesis}  												&cGAN 						&L\ladv, \limage					&--				&M\mae, \psnr					&[\yes] Chest \\
		\cite{ben2018cross} 													&FCN+cGAN 					&L\ladv, \limage					&--				&M\mae, \psnr, \lesiondetection		&[\yes] Liver  \\\midrule
		 \textbf{\textit{PET  $\rightarrow$ CT}} 				&&&&\\
		 \midrule
		\cite{armanious2018medgan} 											& cGAN*						&L\ladv, \limage, \lpercep, \lstylecontent	& --				&M\mae, \psnr, \ssim, \vif, \uqi, \lpips  &[\yes] Brain \\\midrule
		 \textbf{\textit{MR  $\rightarrow$ PET 	}}			&&&&\\
		 \midrule
		\cite{wei2018learning} 												& cascade cGAN				&L\ladv, \limage					& --				&M\tss 						&[\yes] Brain\\
		\cite{pan2018synthesizing} 											& 3D CycleGAN				&L\ladv, \limage, \lcycle				&D\adni 			&M\classification 				&[\yes] Brain\\		
		\midrule
		 \textbf{\textit{PET  $\rightarrow$ MR 	}}			&&&&\\
		 \midrule
		\cite{choi2017generation} 												& pix2pix						&L\ladv, \limage					&D\adni 			&M\ssim, \tss					&[\yes] Brain \\\midrule
		 \textbf{\textit{Synthetic  	$\rightarrow$ Real }}			&&&&\\
		 \midrule
		\cite{hou2017unsupervised} 											&synthesizer+cGAN				&L\ladv, \limage, \lshape 				&D\cbtc, \miccaiseven&M\humanobserver, \segmentation	&[\yes] Histopathology  \\\midrule
		 \textbf{\textit{Real  	$\rightarrow$ Synthetic }}			&&&&\\
		 \midrule
		\cite{mahmood2017unsupervised} 										&cGAN						&L\ladv, \lselfreg					&--				&M\depth	 				&[\no] Endocsocpy \\
		\cite{zhang2018task} 												& CycleGAN*					&L\ladv, \lcycle, \lshape				&--				&M\segmentation 				&[\no	] X-ray \\\midrule
		 \textbf{Domain adaption	}				&&&&\\
		\cite{chen2018semantic} 												& CycleGAN*					&L\ladv, \lcycle, \lshape				&D\montgomery, \jsrt	&M\segmentation 				&[\no	] X-ray \\\midrule
		 \textbf{\textit{T1 $\leftrightarrow$ T2 MR	}}		&&&&\\
		 \midrule
		 \cite{dar2018image} 												&CycleGAN					&L\ladv, \lcycle 						&D\ixi, \bratsfive, \midas &M\psnr, \ssim				&[\no] Brain\\
		\cite{yang2018mri} 													& cGAN						&L\ladv, \limage					&D\bratsfive 		&M\mae, \psnr, \mi, \segmentation, \cmr &[\no] Brain	 \\
		\cite{welander2018generative} 											& CycleGAN, UNIT				&L\ladv, \limage, \lcycle 				&D\humanc			&M\mae, \psnr, \mi  &[\no] Brain	\\
		\cite{liu2018susan} 													& CycleGAN					&L\ladv, \limage, \lcycle 				&D\ski				&M\segmentation   &[\no] Knee	\\		
		\midrule
		 \textbf{\textit{T1 $\rightarrow$ FLAIR MR 	}}		& &&&\\
		 \midrule
		 \cite{yu20183d} 													&cGAN 						&L\ladv, \limage 					&D\bratsfive		&M\mae, \psnr, \segmentation		&[\yes] [3D] Brain	 \\\midrule
		  \textbf{\textit{T1, T2 $\rightarrow$ MRA}}			&&&&&\\
		  \midrule
		 \cite{olut2018generative}												& pix2pix* 					&L\ladv, \limage, \lsteer 				&D\ixi			&M\psnr, \segmentation 			&[\yes]  Brain \\\midrule
		  \textbf{\textit{3T $\rightarrow$ 7T MR }	}			&&&&\\
		  \midrule
		\cite{nie2018medical} 												& Cascade GAN 				&L\ladv, \limage, \lgrad				&  -- 				&M\mae, \psnr 					&[\yes] Brain  \\	\midrule	
		 \multicolumn{6} {@{}l} {\textbf{\textit{Histopathology color normalization}}}\\
		 \midrule
		 \cite{bentaieb2018adversarial}											& cGAN+classifier				&L\ladv, \ledge, \lclassify 				&D\mitos, \glas, \ochd  &M\classification 				&[\no]  \\
		 \cite{zanjani2018histopathology}										&InfoGAN						&L\ladv, \limage, \lselfreg, \lkl 			& -- 				&M\medianintensity				&[\no] \\
		\cite{shaban2018staingan}											&CycleGAN					&L\ladv, \limage, \lcycle 				&D\mitos, \camelyon  	&M\psnr, \ssim, \fsim, \classification &[\no	] \\\midrule
		 \multicolumn{6} {@{}l} {\textbf{\textit{Hyperspectral histology $\rightarrow$ H\&E}}}\\
		 \midrule
		 \cite{bayramoglu2017towards} 											& cGAN						&L\ladv, \limage					&D\histstain			&M\psnr, \ssim  			&[\yes] Lung \\\bottomrule		
\end{tabular}
}


\label{table:cross}

\end{table*}

\subsubsection{Other conditional synthesis}
Medical images can be generated by constraints on segmentation maps, text, locations or synthetic images etc. This is useful to synthesize images in uncommon conditions, such as lung nodules touching the lung border~\citep{jin2018ct}. Moreover, the conditioned segmentation maps can also be generated from GANs~\citep{guibas2017synthetic} or from a pretrained segmentation network~\citep{costa2017towards}, by making the generation a two stage process. ~\cite{mok2018learning} used cGAN to augment training images for brain tumour segmentation. The generator was conditioned on a segmentation map and generated brain MR images in a coarse to fine manner. To ensure the tumour was well delineated with a clear boundary in the generated image, they further forced the generator to output the tumour boundaries in the generation process. 
The full list of synthesis works is summarized in Table~\ref{table:othersyn}.

\begin{table*}[!tp]
\centering
\tiny
\caption{Other conditional image synthesis publications categorized by imaging modality.  * following the method denotes some modifications on the basic framework either on the network architecture or on the employed losses.  A brief description of the losses, quantitative evaluation measures and datasets can be found in Tables ~\ref{table:loss}, ~\ref{table:metric} and~\ref{table:dataset}}.

\resizebox{\textwidth}{!}{
\begin{tabular}{@{}lllll@{}}
	\toprule
		\textbf{Publications} & \textbf{Conditional information}	&  \textbf{Method}  &\textbf{Dataset} &\textbf{Evaluation} \\\midrule
		\textbf{\textit{CT}} &&&&\\
		\cite{jin2018ct} (lung nodule)				& VOI with removed central region  		&[3D] pix2pix* ($\ML_1$ loss considering nodule context)&D\ldct			&M\segmentation\\\midrule
		\textbf{\textit{MR}} &&&&\\
		\midrule
		\cite{mok2018learning}					& Segmentation map  				& Coarse-to-fine boundary-aware&D\bratsfive		&M\segmentation\\
		\cite{shin2018medical}					& Segmentation map  				& pix2pix 				&D\adni, \bratsfive, \bratsseven		&M\segmentation\\
		\cite{gu2018generating}					& MR  							& CycleGAN 			&D\humanc				&M\ssim, \cc \\
		\cite{lau2018scargan}					& Segmentation map  				& Cascade cGAN 		&-						&M\segmentation \\
		\cite{hu2018prostategan}					& Gleason score  					& cGAN 				&-						&- \\
		\midrule
		\textbf{\textit{Ultrasound}} &&&&\\
		\midrule
		\cite{hu2017freehand}  (fetus) 				&  Probe location					& cGAN 				& --					&M\humanobserver\\
		\cite{tom2018simulating}   				&  Segmentation map				& cascade cGAN 		&D\ivus				&M\humanobserver\\ \midrule
		\textbf{\textit{Retinal fundus imaging}}&&&&\\
		\midrule
		\cite{zhao2017synthesizing} 				&  Vessel map  						& cGAN				&D\histstain, \drive, \hrf  	&M\segmentation\\
		\cite{guibas2017synthetic} 				& Vessel map  						& Dual cGAN			&D\drive				&M\qv, \segmentation\\
		\cite{costa2017towards}					&Vessel map  						& Segmentor+pix2pix	&D\drive				&M\qv, \isc\\
		\cite{ costa2017end} 						&Vessel map  						& Adversarial VAE+cGAN	&D\drive, \messidor 		&M\isc \\
		\cite{appan2018retinal}  					& Vessel map; Lesion map  			& cGAN				&D\messidor, \dridb, \diaretdb &M\qv, \lesiondetection \\
		\cite{iqbal2018generative}  				& Vessel map  						& cGAN				&D\drive, \stare 		&M\segmentation \\
		\midrule
		\textbf{\textit{Histopathology}}&&&&\\
		\midrule
		\cite{senaras2018optimized} 				& Segmentation map  				& pix2pix				& --					&M\humanobserver\\\midrule
		\textbf{\textit{X-ray}}&&&&\\
		\midrule
		\cite{galbusera2018exploring}				& Different view; segmentation map 		& pix2pix/CycleGAN		&--					&-- \\
		\cite{mahapatra2018efficient}				& segmentation map+X-ray 			& pix2pix*	 (content loss encourage dissimilarity)	&D\jsrt				&M\classification, \segmentation \\	
		\cite{oh2018learning}					& X-ray (for bone supression) 			& pix2pix* (Haar wavelet decomposion)	 		&--					&M\mae, \psnr, \ssim, \nps \\		
	
		\bottomrule
\end{tabular}  
}
\label{table:othersyn}
\end{table*}

\begin{table*}[!htb]
\centering
\tiny
\caption{Common datasets used in the reviewed literature. In the first column, D stands for Dataset.}
\resizebox{\textwidth}{!}{
\begin{tabular}{@{}lllll@{}}
	\toprule
		\textbf{Abbre.} 					&	\textbf{Dataset}																			&	\textbf{Purpose}						&\textbf{Anatomy}				&\textbf{Modality}	\\\midrule
		D\piglet~~\cite{yi2018sharpness}	&	 \href{https://github.com/xinario/SAGAN}{Piglet}  												& Denoising								&Whole body 					& CT\\
		D\ldct~~\cite{ldct2017} 			&	\href{https://www.aapm.org/GrandChallenge/LowDoseCT/}{LDCT2016}								&Denoising								& Abdomen 					&CT\\
		D\miccaigcthree 				&	\href{https://my.vanderbilt.edu/masi/workshops/}{MICCAI2013} 										& Organ segmentation						& Abdomen, Pelvis 				& CT \\
		D\lidc~\cite{LIDC}				&	 \href{https://wiki.cancerimagingarchive.net/display/Public/LIDC-IDRI}{LIDC-IDRI} 						&Lung cancer detection and diagnosis			& Lung						& CT\\
		D\deeplesion~\cite{yan2018deep}	&	 \href{https://wiki.cancerimagingarchive.net/display/Public/LIDC-IDRI}{DeepLesion} 						&Lesion segmentation						& --							& CT\\		
		D\lits							&	 \href{https://competitions.codalab.org/competitions/17094}{LiTS2017} 								&Liver tumor segmentation					& Liver						& CT\\
		D\spinebiomedia~\cite{glocker2013vertebrae}	&	 \href{https://biomedia.doc.ic.ac.uk/data/spine/}{Spine} 									&Vertebrate localization						& Spine						& CT\\
		D\nsclc~\cite{nsclc}				&	 \href{https://wiki.cancerimagingarchive.net/display/Public/NSCLC-Radiomics\#23aa293d4c344848ade0eaadfdcf87dd}{NSCLC-Radiomics} 					&Radiomics			& Lung						& CT\\
		
		D\wholeheart~\cite{zhuang2016multi}&	\href{http://www.sdspeople.fudan.edu.cn/zhuangxiahai/0/mmwhs/}{MM-WHS}							&Whole heart segmentation					& Heart						&CT, MR\\
		D\wholeheartvessel~\cite{pace2015interactive}&	\href{http://www.sdspeople.fudan.edu.cn/zhuangxiahai/0/mmwhs/}{HVSMR 2016}				&Whole heart and great vessel segmentation		& Heart, Vessel						&MR\\
		D\ixi							&	 \href{http://brain-development.org/ixi-dataset/}{IXI}												&Analysis of brain development 				& Brain						& MR\\
		D\dsbonefive					&	 \href{https://www.kaggle.com/c/second-annual-data-science-bowl/data}{DSB2015} 						&End-systolic/diastolic volumes measurement 		&Heart						&MR \\
		D\mridata						&	\href{ http://mridata.org}{Mridata}															& MRI reconstruction						&Knee						& MR\\
		D\ski							&	\href{ http://www.ski10.org}{Ski10}															& Cartilage and bone segmentation				&Knee						& MR\\
		D\brainles~~\cite{brainles}			&	\href{http://www.brainlesion-workshop.org}{BrainLes}	 											&Lesion segmentation						&Brain						&MR\\
		D\adni						&	 \href{http://adni.loni.usc.edu}{ADNI} 															& Alzheimer’s disease neuroimaging Initiative 		&Brain						& MR, PET\\
		D\miccaionetwo				&	\href{https://my.vanderbilt.edu/masi/workshops/}{MAL}\						&Brain structure segmentation					& Brain						& MR\\
		D\bratsthree					&	\href{https://www.smir.ch/BRATS/Start2013}{BRATS2013}\										&Gliomas segmentation						& Brain						& MR\\
		D\bratsfive					&	\href{https://www.smir.ch/BRATS/Start2015}{BRATS2015}\										&Gliomas segmentation						& Brain						& MR\\
		D\bratssix						&	\href{https://sites.google.com/site/braintumorsegmentation/home/brats_2016}{BRATS2016} 				&Gliomas segmentation						& Brain						& MR\\
		D\bratsseven					&	 \href{https://www.med.upenn.edu/sbia/brats2017.html}{BRATS2017}								&Gliomas segmentation, overall survival prediction	& Brain						& MR\\
		D\midas~\cite{midas}			&	\href{http://insight-journal.org/midas/community/view/21}{MIDAS}									& Assessing the effects of healthy aging			& Brain						& MR\\
		D\blsa~\cite{resnick2003longitudinal}&	\href{https://www.blsa.nih.gov}{BLSA}														& Baltimore longitudinal study of aging			& Brain						& MR\\
		D\humanc	~\cite{HCP}			&	 \href{http://www.humanconnectomeproject.org}{HCP} 											&Human connectome project					&Brain						&MR\\
		D\iseg	~\cite{iseg2017}			&\href{http://iseg2017.web.unc.edu}{iSeg2017} 														&Infant brain tissue segmentation			&Brain						&MR\\
		D\biobank						&\href{https://www.ukbiobank.ac.uk}{UK Biobank} 														&Health research						&Brain, Heart, Body						&MR\\
		D\isicsix~\cite{ISIC2016}			&	\href{https://challenge.kitware.com/\#challenge/560d7856cad3a57cfde481ba}{ISIC2016}					&Skin lesion analysis						&Skin 						&Dermoscopy\\
		D\isicseven~\cite{ISIC2017}		&	 \href{https://challenge.kitware.com/\#challenge/583f126bcad3a51cc66c8d9a}{ISIC2017}					&Skin lesion analysis						&Skin 						&Dermoscopy\\
		D\isiceight						&	\href{https://challenge2018.isic-archive.com}{ISIC2018}											&Skin lesion analysis						&Skin 						&Dermoscopy\\
		D\phtwo~\cite{PH2}				&	\href{http://www.fc.up.pt/addi/ph2\%20database.html}{PH2}										&Skin lesion analysis						&Skin 						&Dermoscopy\\
		D\dermofit~\cite{ballerini2013color}	&	\href{https://licensing.eri.ed.ac.uk/i/software/dermofit-image-library.html}{Dermofit}						&Skin lesion analysis						&Skin 						&Dermoscopy\\
		D\montgomery~\cite{Montgomery}	&  \href{https://ceb.nlm.nih.gov/repositories/tuberculosis-chest-x-ray-image-data-sets/}{Montgomery}				&Pulmonary disease detection					&Chest						&X-Ray\\
		D\jsrt	~\cite{JSRT}				&\href{http://db.jsrt.or.jp/eng.php}{JSRT}															&Pulmonary nodule detection					&Chest						& X-Ray\\	
		D\plco						&	\href{https://biometry.nci.nih.gov/cdas/plco/}{NIH PLCO} 											&\makecell[l]{Cancer screening trial for \\Prostate, lung, colorectal and ovarian (PLCO)} &- &X-ray; Digital pathology\\
		D\cbtc						&\href{https://wiki.cancerimagingarchive.net/pages/viewpage.action?pageId=20644646}{CBTC2015} 			& Segmentation of nuclei						& Nuclei						& Digital pathology\\
		D\miccaiseven					&\href{http://miccai.cloudapp.net/competitions/57}{CPM2017}  											& Segmentation of nuclei						& Nuclei						& Digital pathology\\
		D\mitos						&	\href{https://mitos-atypia-14.grand-challenge.org}{MITOS-ATYPIA}									& Mitosis detection; Nuclear atypia score evaluation &Breast						&Digital pathology\\
		D\glas~\cite{GLAS}				&	\href{https://warwick.ac.uk/fac/sci/dcs/research/tia/glascontest/}{GlaS}								&Gland segmentation						&Colon						&Digital pathology\\
		D\ochd~\cite{OCHD}				&	\href{http://ensc-mica-www02.ensc.sfu.ca/download/}{OCHD}										&Carcinoma subtype prediction					&Ovary						&Digital pathology\\
		D\camelyon					&	 \href{https://camelyon16.grand-challenge.org/}{Camelyon16}										&Lymph node metastases detection				&Breast						&Digital pathology\\
		D\histstain	~\cite{VirtualStaining}	&	 \href{http://www.ee.oulu.fi/~nyalcinb/papers/iccv2017/index.html}{Neslihan}							&Virtual H\&E staining 						&Lung						&Digital pathology\\
		D\celldetect~\cite{kainz2015you}	&	 \href{https://github.com/pkainz/MICCAI2015/}{CellDetect}											&Cell detection 								&Bone marrow					&Digital pathology\\
		D\drive~\cite{DRIVE}				&	 \href{https://www.isi.uu.nl/Research/Databases/DRIVE/}{DRIVE}									&Blood vessels segmentation					& Eye						&Fundus imaging\\
		D\stare~						&	\href{http://cecas.clemson.edu/~ahoover/stare/}{STARE}											&Structural analysis of the retina				&Eye						&Fundus Imaging\\
		D\hrf~\cite{HRF}				&	\href{https://www5.cs.fau.de/research/data/fundus-images/}{HRF}									&Image quality assessment, segmentation		&Eye						&Fundus Imaging\\
		D\messidor~\cite{Messidor}		&	 \href{http://www.adcis.net/en/Download-Third-Party/Messidor.html}{Messidor}							&Segmentation in retinal ophthalmology 			&Eye						&Fundus Imaging\\
		D\dridb~\cite{DRiDB}			& \href{https://ipg.fer.hr/ipg/resources/image_database}{DRiDB}										&Diabetic retinopathy detection				& Eye						&Fundus Imaging\\
		D\diaretdb	~\cite{diaretdb1}		&	\href{http://www.it.lut.fi/project/imageret/diaretdb1/}{DIARETDB1}									&Diabetic retinopathy detection				& Eye						&Fundus Imaging\\
		D\rimone	~\cite{fumero2011rim}	&	RIM-ONE																				&Optic nerve head segmentation				& Eye						&Fundus Imaging\\
		D\ithreea	~\cite{hobson2015benchmarking}&	\href{http://nerone.diem.unisa.it/hep2-benchmarking/dbtools/}{I3A}								&HEp-2 cell classification						& Skin						&Fluorescent microscopy \\
		D\mivia						&	\href{https://mivia.unisa.it}{MIVIA}														&HEp-2 cell segmentation						& Skin						&Fluorescent microscopy \\
		D\ivus~\cite{ivus}				&	\href{http://www.cvc.uab.es/IVUSchallenge2011/dataset.html}{IVUS}									&Vessel inner and outer wall border detection		& Blood Vessel					&Ultrasound\\
		D\inbreast~\cite{moreira2012inbreast}	&\href{http://medicalresearch.inescporto.pt/breastresearch/index.php/Get_INbreast_Database}{INbreast}		&Mass segmentation							&Breast						& Mammography\\	
		D\ddsm~\cite{heath1998current}		&\href{http://www.eng.usf.edu/cvprg/Mammography/Database.html}{DDSM-BCRP}						&Mass segmentation							&Breast						& Mammography\\	

		\bottomrule
\end{tabular}
}
\label{table:dataset}
\end{table*}

\renewcommand\arraystretch{0.8}

\begin{table*}[!tp]
\centering
\tiny
\caption{Segmentation publications. A brief description of the  datasets can be found in Table~\ref{table:dataset}.}
\resizebox{\textwidth}{!}{
\begin{tabulary}{\textwidth}{@{}llp{0.7\textwidth}@{}}
	\toprule
		\textbf{Publications}		&\textbf{Dataset}   	& \textbf{Remarks}\\\midrule
		\textbf{\textit{CT}}				&&\\
		\midrule
		\cite{yang2017automatic}			&-- 					&[3D] [Liver] Generator is essentially a U-net with deep supervisions\\
		\cite{dou2018unsupervised}		&D\wholeheart			& Ensure that the feature distribution of images from both domains (MR and CT) are indistinguishable\\
		\cite{rezaei2018conditional}		&D\lits				& Additional refinement network, patient-wise batchNorm, recurrent cGAN to ensure temporal consistancy \\
		\cite{sekuboyina2018btrfly}		&D\spinebiomedia		& Adversarial training based on EBGAN; Butterfly shape network to combine two views \\
		 \midrule
		\textbf{\textit{MR}}				&&\\
		\midrule
		\cite{xue2018segan}  			&D\bratsthree, \bratsfive		&A multi-scale $L_1$ loss in  the discriminator where features coming from different depth are compared	\\
		\cite{rezaei2017conditional}		&D\bratsseven				& The generator takes heterogenous MR scans of various contrast as provided by BRATS 17 challenge\\
		\cite{rezaei2018whole}			&D\wholeheartvessel			&A cascade of cGANs in segmenting myocardium and blood pool\\
		\cite{li2017brain}				&D\bratsseven				& The generator takes heterogenous MR scans of various contrast as provided by BRATS 17 challenge\\
		\cite{moeskops2017adversarial}	&D\miccaionetwo, \bratsthree	&--				\\
		\cite{kohl2017adversarial}			&--						&[Prostate] Improved sensitivity\\
		\cite{huo2018splenomegaly}		&-- 						&[Spleen] Global convolutional network (GCN) with a  large receptive field as the generator\\
		\cite{kamnitsas2017unsupervised}	&--						& Regulate the learned representation so that the feature representation is domain  invariant\\
		\cite{dou2018unsupervised}		&D\wholeheart				& Ensure that the feature distribution of images from both domains (MR and CT) are indistinguishable\\
		\cite{rezaei2018conditional}		&D\bratsseven				& Additional refinement network, patient-wise batchNorm, recurrent cGAN to ensure temporal consistency\\
		\cite{xu2018mutgan}				&--						& Joint learning (segmentation and quantification); convLSTM in the generator for spatial-temporal processing; Bi-LSTM in the discriminator to learn relation between tasks \\	
		\cite{han2018spine}				&--						&Local-LSTM in the generator to capture spatial  correlations between neighbouring structures \\
		\cite{zhao2018craniomaxillofacial}	&D\adni					&Deep supervision; Discriminate segmentation map based on features extracted from a pre-trained network \\		
		 \midrule
		\textbf{\textit{Retinal fundus imaging}}	&&\\
		\midrule
		\cite{son2017retinal}				&D\drive, \stare				&Deep architecture is better for discriminating whole images and has less false positives with fine vessels\\
		\cite{zhang2017deep}			&D\glas					&Use both annotated and unannotated images in the segmentation pipeline\\
		\cite{shankaranarayana2017joint}	&D\rimone				&--	\\ \midrule
		\textbf{\textit{X-ray}}				&&\\
		\midrule
		\cite{dai2017scan}				&D\montgomery, \jsrt			&Adversarial loss is able to correct the shape inconsistency\\ \midrule
		\textbf{\textit{Histopothology}}		&&\\
		\midrule
		\cite{wang2017adversarial}		&--						&Basal membrane segmentation\\\midrule
		\textbf{\textit{fluorescent microscopy}	}	&&\\
		\midrule
		\cite{li2018cc}					&D\ithreea, \mivia			&pix2pix + ACGAN; Auxiliary classifier branch provides regulation to  both the discriminator and the segmentor\\\midrule
		\textbf{\textit{Dermoscopy}}			&&\\
		\midrule
		\cite{izadi2018generative}			&D\dermofit				&Adversarial training helps to refine the boundary precision\\\midrule
		\textbf{\textit{Mammography}}		&&\\
		\midrule
		\cite{ zhu2017adversarial}			&D\inbreast, \ddsm			&Enforce network invariance to small perturbations of the training samples in order to reduce overfitting on small size dataset\\
		\midrule
		\textbf{\textit{Ultrasound}}		&&\\
		\midrule
		\cite{tuysuzoglu2018deep}	&--							&Joint learning (landmark localization + prostate contour segmentation); Contour shape prior imposed by the discriminator\\
		 \bottomrule		
\end{tabulary}
}
\label{table:segment}
\end{table*}

\subsection{Segmentation}
Generally, researchers have used pixel-wise or voxel-wise loss such as cross entropy for segmentation. Despite the fact that U-net~\citep{ronneberger2015u} was used to combine both  low-level and high-level features, there is no guarantee of spatial consistency in the final segmentation map.  Traditionally, conditional random field (CRF) and graph cut methods are usually adopted for segmentation refinement by incorporating spatial correlation. Their limitation is that they only take into account pair-wise potentials which  might cause serious boundary leakage in low contrast regions. On the other hand, adversarial losses as introduced by the discriminator can take into account high order potentials~\citep{yang2017automatic}. In this case, the discriminator can be regarded as a shape regulator. This regularization effect is more prominent when the object of interest has a compact shape, e.g. for lung and heart mask but less useful for deformable objects such as vessels and catheters.  This regulation effect can be also applied to the internal features of the segmentor to achieve domain (different scanners, imaging protocols, modality) invariance~\citep{kamnitsas2017unsupervised, dou2018unsupervised}. The adversarial loss can also be viewed as a adaptively learned similarity measure between the segmented outputs and the annotated groundtruth. Therefore, instead of measuring the similarity in the pixel domain, the discriminative network projects the input to a low dimension manifold and measures the similarity there. The idea is similar to the perceptual loss. The difference is that the perceptual loss is computed from a pre-trained classification network on natural images whereas the adversarial loss  is computed from a network that trained adaptively during the evolvement of the generator.

~\cite{xue2018segan} used a multi-scale $L_1$ loss in  the discriminator where features coming from different depths are compared. This was demonstrated  to be effective in enforcing the multi-scale spatial constraints on segmentation maps and the system achieved state-of-the-art performance in the BRATS 13 and 15 challenges.   \cite{zhang2017deep} proposed to use both annotated and unannotated images in the segmentation pipeline. The annotated images are used in the same way as in~\citep{xue2018segan, son2017retinal} where both element-wise loss and adversarial loss are applied. The unannotated images on the other hand are only used to compute a segmentation map to confuse the discriminator.  \cite{li2018cc} combined pix2pix with ACGAN for segmentation of fluorescent microscopy images of different cell types. They found that the introduction of the auxiliary classifier branch provides regulation to both the discriminator and the segmentor.

Unlike these aforementioned segmentation works where adversarial training is used to ensure higher order structure consistency on the final segmentation maps, the adversarial training scheme in ~\citep{zhu2017adversarial}  enforces network invariance to small perturbations of the training samples in order to reduce overfitting on small dataset.  
Papers related to medical image segmentation are summarized  in Table~\ref{table:segment}.


\subsection{Classification}
%
%
Classification is arguably  one of the most successful tasks where  deep learning has been applied. Hierarchical image features can be extracted from a  deep neural network discriminatively trained with image-wise class labels.  GANs have been used for classification problems as well, either  using part of the generator and discriminator as a feature extractor or  directly using the discriminator as a classifier (by adding an extra class corresponding to the generated images).  \cite{hu2017unsupervised} used combined WGAN and InfoGAN for unsupervised cell-level feature representation learning in histopathology images whereas ~\cite{yi2018unsupervised} combined WGAN and CatGAN for unsupervised and semi-supervised feature representation learning for dermoscopy images. Both works extract features from the discriminator and build a classifier on top. 
~\cite{madani2018semi}, ~\cite{lahiri2017generative} and~\cite{lecouat2018semi} adopted the semi-supervised training scheme of GAN for chest abnormality classification, patch-based retinal vessel  classification and cardiac disease diagnosis respectively. They   found that the semi-supervised GAN can achieve  performance comparable with a traditional supervised CNN with an order of magnitude less labeled data.  Furthermore, ~\cite{madani2018semi}  have also shown that the adversarial loss can reduce domain overfitting by simply supplying unlabeled test domain images to the discriminator in identifying cardiac abnormalities in chest X-ray. A similar work in addressing domain variance in whole slide images (WSI) has been conducted by~\cite{ren2018adversarial}.

Most of the other works that  used GANs to generate new training samples have been already mentioned  in Section~\ref{sect:us}. These studies applied a two stage process, with the first stage learned to augment the images and the second stage learned to perform classification by adopting the traditional classification network. The two stages are trained disjointedly  without any communication in between. The advantage is that these two components can be replaced easily if more advanced unconditional synthesis architectures are proposed whereas the downside  is that the generation has to be conducted for each class separately (N models for N classes), which is not memory and computation efficient.  A single model that is capable of performing conditional synthesis of multiple categories is an active research direction~\citep{brock2018large}. Surprisingly,~\cite{frid2018gan} found that using separate GAN (DCGAN) for each lesion class resulted in better performance in lesion classification than using a unified GAN (ACGAN) for all classes. The underlying reason remains to be explored.  Furthermore,~\cite{finlayson2018towards} argue that images generated from GANs  may serve as an effective augmentation in the medium-data regime, but may not be helpful in a high or low-data regime.

%
\subsection{Detection}
The discriminator of GANs can be utilized to detect abnormalities such as lesions by learning the probability distribution of training images depicting normal pathology. Any image that falls out of this distribution can be deemed as abnormal.~\cite{schlegl2017unsupervised} used the exact idea to learn a manifold of normal anatomical variability and proposed a novel anomaly scoring scheme based on the fitness of the test image's latent code to the learned manifold. The  learning process was conducted in an unsupervised fashion and  effectiveness was demonstrated by state-of-the-art performance of anomaly detection on optical coherence tomography (OCT) images. 
~\cite{alex2017generative} used GAN for brain lesion detection on MR images. The generator was used to model the distribution of normal patches and the trained discriminator was used to compute a posterior probability of patches centered on every pixel in the test image. \cite{chen2018unsupervised} used an adversarial auto-encoder to learn the data distribution of healthy brain MR images. The lesion image was then mapped to an image without a lesion by exploring the learned latent space, and the lesion could be highlighted by computing the residual of these two images.  We can see that all the detection studies targeted for abnormalities that are hard to enumerate.

In the image reconstruction section, it has been observed that if the target distribution is formed from medical images without pathology, lesions within an image could be removed in the CycleGAN-based unpaired image transfer due to the distribution matching effect. However, it can be seen here that if the target and source domain are of the same imaging modality differing only in terms of normal and abnormal tissue, this adverse effect can actually be exploited for abnormality detection~\cite{sun2018adversarial}.

\subsection{Registration}
 cGAN can also be used for multi-modal or uni-modal image registration. The generator in this case will either generate transformation parameters, e.g. 12 numbers for 3D affine transformation, deformation field for non-rigid transformation or directly generate  the transformed image. The discriminator then discriminates aligned image pairs from unaligned image pairs. A spatial transformation network~\citep{jaderberg2015spatial} or a deformable transformation layer~\citep{fan2018adversarial}  is usually  plugged in between these two networks to enable end-to-end training. \cite{yan2018adversarial} performed prostate MR to transrectal ultrasound (TRUS) image registration using this framework. The paired training data was obtained through manual registration by experts.  ~\cite{yan2018adversarial} employed a discriminator to regularize the displacement field computed by the generator and found this approach to be more effective than the other regularizers in MR to TRUS registration. ~\cite{mahapatra2018deformable} used CycleGAN for multi-modal (retinal) and uni-modal (MR) deformable registration where the generator produces both the transformed image and the deformation field. \cite{mahapatra2018joint} took one step further and explored the idea of joint segmentation and registration with CycleGAN and found their method performs better than the separate approaches for lung X-ray images. ~\cite{tanner2018generative} employed CycleGAN for deformable image registration between MR and CT by first transforming the source domain image to the target domain and then employing  a mono-modal image similarity measure for the registration. They found this method can  achieve at best similar performance with the traditional multi-modal deformable registration methods. 

\subsection{Other works}
In addition to the tasks described in the aforementioned sections, GANs have also been applied in other tasks discussed here. For instance, cGAN has been used for modelling patient specific motion distribution based on a single preoperative image \citep{hu2017intraoperative}, highlighting regions most accountable for a disease ~\citep{baumgartner2017visual} and re-colorization of endoscopic video data~\citep{ross2018exploiting}. In ~\citep{mahmood2018automated} pix2pix was used for treatment planning in radiotherapy by predicting the dose distribution map from CT image. WGAN has also been used for modelling the progression of Alzheimer's disease (AD) in MRI. This is achieved by isolating the latent encoding of AD and performing arithmetic operation in the latent space ~\citep{bowles2018modelling}.

\section{Discussion}\label{sect:discuss}
In the years 2017 and 2018, the number of studies applying GANs has risen significantly. The list of these papers reviewed for our study can be found on our~\footnote{https://github.com/xinario/awesome-gan-for-medical-imaging}{GitHub repository}. 

About 46\% of these papers studied  image synthesis, with cross modality image synthesis being the most important application of GANs. MR is ranked as the most common imaging modality explored in the GAN related literature. We believe one of the reasons for the significant interest in applying GANs for MR image analysis is due to the excessive amount of  time  spent on the acquisition of multiple sequences. GANs hold  the potential to reduce MR acquisition time by faithfully generating certain sequences from already acquired ones. A recent study in image synthesis across different MR sequences using CollaGAN  shows the irreplaceable nature of exogenous contrast sequence, but reports the synthesis of endogenous contrast such as T1, T2, from each other with high fidelity~\citep{lee2019contrast}.  A second reason for the popularity of GANs in MR might be because of large number of publicly available MR datasets  as shown in Table~\ref{table:dataset}.  

Another 37\% of these studies fall into the group of reconstruction and segmentation  due to the popularity of image-to-image translation frameworks. Adversarial training in these cases imposes a strong shape and texture regulation on the generator's output which makes it very promising in these two tasks.  For example, in  liver segmentation from 3D CT volumes, the incorporation of adversarial loss significantly improves the segmentation performance on non-contrast CT (has fuzzy liver boundary) than graph cut and CRF~\citep{yang2017automatic}.  

Further 8\% of these studies are related to classification. In these studies, the most effective use case was to combat domain shift.  For the studies that used GAN for data augmentation in classification, most focused on generating tiny objects that can be easily aligned, such as nodules, lesions and cells. We believe it is partly due to the relatively smaller content variation of these images compared to the full context image which makes the training more stable with the current technique. Another reason might be related to the computation budget of the research since training on high resolution images requires a lot of GPU time.   Although there are studies that applied GAN on synthesizing whole chest-X-ray~\citep{madani2018chest, madani2018semi}, the effectiveness has only been shown on fairly easy tasks, e.g. cardiac abnormality classification and on a medium size data regime, e.g. a couple of thousand images. With the advent of  large volume labeled datasets, such as the CheXpert~\citep{irvin2019chexpert}, it seems there is diminishing return in the employment of GANs for image generation, especially for classification. We would like to argue that  GANs are still useful in the following two cases. First, nowadays the training of a deep neural network heavily relies on data augmentation to improve the network's generalizability on unseen test data and reduce overfitting. However, existing data augmentation operations are all manually designed operations, e.g. rotation, color jittering, and can not cover the whole variation of the data. ~\cite{cubuk2018autoaugment} recently  proposed to learn an augmentation policy with reinforcement learning but the search space  still consisted of basic hand-crafted image processing operations. GANs, however,  can allow us to sample the whole data distribution which offers much more flexibility in augmenting the training data~\citep{bowles2018gan}. For example, styleGAN, is able to generate high resolution realistic face images with unprecedented level of details. This could be readily applied to chest X-ray datasets to generate images of a  pathology class that has sufficient number of cases. Second, it is well known that medical data distribution is highly skewed with its largest mass centered on common diseases. It is impossible to accumulate enough training data for rare diseases, such as rheumatoid arthritis, sickle cell disease. But radiologists have been trained to detect these diseases in the long tail. Thus, another  potential of GANs will be in synthesizing uncommon pathology cases, most
likely through conditional generation with the conditioned information being specified by medical experts either through text description or hand drawn figures.  

The remaining studies pertaining to detection, registration and other applications are so limited that it is hard to draw any conclusion.


\subsection{Future challenges}
Alongside many positive utilities of GANs,  there are still challenges that need to be resolved for their employment to medical imaging.  In image reconstruction and cross modality image synthesis, most works still adopt traditional shallow reference metrics such as MAE, PSNR, or SSIM for quantitative evaluation. These measures, however, do not correspond to the visual quality of the image. For example, direct optimization of pixel-wise loss produces a suboptimal (blurry) result but provides higher numbers  than using adversarial loss. It becomes increasingly difficult to interpret these numbers in horizontal comparison of GAN-based works especially when extra losses as shown in Table~\ref{table:loss} are incorporated.  One way to alleviate this problem is to use down stream tasks such as segmentation or classification to validate the quality of the generated sample. Another way is to recruit domain experts but this approach is expensive, time consuming and hard to scale. Recently, \cite{zhang2018unreasonable} proposed learned perceptual image path similarity (LPIPS), which outperforms previous metrics in terms of agreement with human judgements. It has been adopted in MedGAN~\citep{armanious2018medgan} for evaluation of the generated image quality but it would be interesting to see its effectiveness for different types of medical images as compared to subjective measures from experienced human observers in a more extensive study. For natural images, the unconditional generated sample quality and diversity is usually measured by inception score~\citep{salimans2016improved}, the mean MS-SSIM metric among randomly chosen synthetic sample pairs~\citep{odena2016conditional}, or Fr\'echet Inception distance (FID)~\citep{heusel2017gans}. The validity of these metrics for medical images remains to be explored.

Cross domain image-to-image translation can be achieved with both paired and unpaired training data and it  offers many prospective applications in medical imaging as has already been seen in section~\ref{crossmodsyn}. Unpaired training does not have the data fidelity loss term therefore there is no guarantee of preservation of small abnormality regions during the translation process. ~\cite{cohen2018distribution} warn against the use of generated images for direct interpretation by doctors. They observe that trained CycleGAN networks (for unpaired data) can be subject to bias due to matching the generated data  to the distribution of the target domain. This system bias comes into being when  target domain images in the  training set  have an over or under representation of certain classes.  As an example of exploitation of this effect,~\cite{mirsky2019ct} demonstrate  the possibility of malicious tampering of 3D medical imaging using 3D conditional GANs to remove and inject solitary pulmonary nodule into patient's CT scans. This system bias also exists in paired cross domain image-to-image translation with the data fidelity loss but only happens when the model was trained on normal images but tested on abnormal images.   Cautions should be taken in training of the translation model and new methods should be proposed to faithfully preserve local abnormal regions.

\subsection{Interesting future applications}

Similar to other deep learning neural network models, various applications of GANs demonstrated in this paper have direct bearing on improving radiology workflow and patient care. The strength of GANs however lies in their ability to learn in an unsupervised and/or weakly-supervised fashion. In particular, we perceive that image-to-image translation achieved by cGANs can have various other useful applications in medical imaging. For example, restoration of MR images acquired with certain artifacts such as motion, especially in a pediatric setting, may help reduce the number of repeated exams. 

Exploring GANs for image captioning task~\citep{dai2017towards, shetty2017speaking, melnyk2018improved, fedus2018maskgan} may lead to semi-automatic generation of medical imaging reports~\citep{jing2017automatic} potentially reducing image reporting times. Success of adversarial text classification~\citep{liu2017adversarial} also prompts potential utility of GANs in improving performance of such systems for automatic MR protocol generation from free-text clinical indications~\citep{jae2017}. Automated systems may improve MRI wait times which have been on the rise~\citep{wait2017} as well as enhance patient care. cGANs, specifically CycleGAN applications, such as makeup removal~\citep{chang2018pairedCycleGAN}, can be extended to medical imaging with applications in improving bone x-ray images by removal of artifacts such as casts to facilitate enhanced viewing. This may aid radiologists in assessing fine bony detail, potentially allowing for enhanced detection of initially occult fractures and helping assess the progress of bone healing more efficiently.  The success of GANs in unsupervised anomaly detection~\citep{schlegl2017unsupervised} can help achieve the task of detecting abnormalities in medical images in an unsupervised manner.  This has the potential to be further extended for detection of  implanted devices, e.g. staples, wires, tubes, pacemaker and artificial valves on X-rays. Such an algorithm can also be used for prioritizing radiologists' work lists, thus reducing the turnaround time for reporting critical findings~\citep{gal2018}.   
We also expect to witness the utility of GANs in medical image synthesis from text descriptions~\citep{bodnar2018text}, especially for rare cases, so as to fill in the gap of training samples required for training supervised neural networks for medical image classification tasks. The recent work on styleGAN shows the capability to control ~\citep{karras2018style}   the high level attributes of the synthesized image by manipulating the scale and bias parameters of the AdaIN layer~\citep{huang2017arbitrary}. Similarly, the SPADE~\citep{park2019semantic}  controls the semantic layout of the synthesized image by a spatially adaptive normalization layer. Imagine in the future the desired attribute can be customized and specified in prior and manipulated in a localized fashion. We may then be able to predict the progression of disease, measure the impact of drug trial as suggested in~\cite{bowles2018modelling} but with more fine-grained controls.

Different imaging modalities work by exploiting tissue response to a certain physical media, such as x-rays or a magnetic field, and thus can provide complementary diagnostic information to each other. As a common practice in supervised deep learning, images of one modality type are labelled to train a network to accomplish a desired task. This process is repeated when switching modalities even if the underlying anatomical structure is the same, resulting in a waste of human effort. Adversarial training, or more specifically unpaired cross modality translation, enables reuse of the labels in all modalities and opens new ways for unsupervised transfer learning~\citep{dou2018unsupervised, ying2019x2ct}. 

Finally, we would like to point out that, although there have many promising results reported in the literature, the adoption of GANs in medical imaging is still in its infancy and there is currently no breakthrough application as yet adopted clinically for  GANs-based methods.

\bibliographystyle{model2-names}\biboptions{authoryear}
\bibliography{gan_medical_imaging_final,datasetreferences.bib}

\end{document}